\documentclass[final]{cvpr}

\usepackage{times}
\usepackage{epsfig}
\usepackage{graphicx}
\usepackage{amsmath}
\usepackage{amssymb}
\usepackage{subfigure}
\usepackage{amsthm}

\usepackage{multirow}
\usepackage{booktabs}
\usepackage{algorithm}
\usepackage{algorithmic}
\usepackage{gensymb} % degree symbol
\usepackage{enumitem} % set itemize
\usepackage{color,soul}
\usepackage{mathrsfs}

\usepackage{multirow}
\usepackage{tabularx} % allow set table width
\usepackage{booktabs}
\usepackage{filecontents}

\newcolumntype{L}[1]{>{\hsize=#1\hsize\raggedright\arraybackslash}X}%
\newcolumntype{R}[1]{>{\hsize=#1\hsize\raggedleft\arraybackslash}X}%
\newcolumntype{C}[1]{>{\hsize=#1\hsize\centering\arraybackslash}X}%

\graphicspath{{figure/}}

\usepackage[pagebackref=true,breaklinks=true,colorlinks,bookmarks=false]{hyperref}

\DeclareMathOperator{\tr}{tr}

\newcommand{\cE}{\mathcal{E}}
\newcommand{\cG}{\mathcal{G}}
\newcommand{\cV}{\mathcal{V}}

\newtheorem{theorem}{Theorem}[section]
\newtheorem{lemma}[theorem]{Lemma}
\newtheorem{corollary}{Corollary}[theorem]

\def\cvprPaperID{2419} 
\def\confYear{CVPR 2021}

\begin{document}

%%%%%%%%% TITLE
\title{Rotation Coordinate Descent for Fast Globally Optimal Rotation Averaging}
\author{Álvaro Parra$^{1\thanks{equal contribution}}$ \hspace{1em} Shin-Fang Chng$^{1 \footnotemark[1]}$ \hspace{1em} Tat-Jun Chin$^{1}$ \hspace{1em} Anders Eriksson$^{2}$ \hspace{1em}  Ian Reid$^{1}$  \\
$^{1}$ School of Computer Science, The University of Adelaide \\
$^{2}$ School of Information Technology and Electrical Engineering, University of Queensland}
%The University of Adelaide\\
%\and
%Shin-Fang Chng$^{*}$\\

\maketitle

%%%%%%%%% ABSTRACT
\begin{abstract}

%Under mild conditions on the noise level of the measurements, rotation averaging satisfies strong duality, which enables global solutions to be obtained via semidefinite programming (SDP) relaxation. However, generic solvers for SDP are rather slow in practice, even on instances of moderate size, thus developing specialised algorithms for rotation averaging is vital. In this paper, we present a fast algorithm that achieves global optimality which resembles the block-coordinate-descentent (BCD) method to solve SDP. However, our algorithm does not require iterating over semidefinite matrices as BCD does with updates in a row-by-row fashion. Instead, our algorithm updates estimates for all rotations at each iteration--contrast to BCD for which elements in the rows of its positive semidefinite matrix converge to rotations\footnote{Precisely to orthogonal matrices.  The optimal rotations comes from choosing their signs such that their determinant are positive.}. We theoretically demonstrate the convergence of our algorithm and empirically show superior efficiency than the state-of-the-art global methods over a variety of problem configurations. Solving for rotations at every iteration facilitates the incorporation of local optimisation for further speed-ups. Moreover, our algorithm is trivial to implement\footnote{See supplementary material for our implementation and a demonstration.}. 

Under mild conditions on the noise level of the measurements, rotation averaging satisfies strong duality, which enables global solutions to be obtained via semidefinite programming (SDP) relaxation. However, generic solvers for SDP are rather slow in practice, even on rotation averaging instances of moderate size, thus developing specialised algorithms is vital. In this paper, we present a fast algorithm that achieves global optimality called rotation coordinate descent (RCD). Unlike block coordinate descent (BCD) which solves SDP by updating the semidefinite matrix in a row-by-row fashion, RCD directly maintains and updates all valid rotations  throughout the iterations. This obviates the need to store a large dense semidefinite matrix. We mathematically prove the convergence of our algorithm and empirically show its superior efficiency over state-of-the-art global methods on a variety of problem configurations. Maintaining valid rotations also facilitates incorporating local optimisation routines for further speed-ups. Moreover, our algorithm is simple to implement~\footnote{Source code is available at \url{https://github.com/sfchng/Rotation\_Coordinate\_Descent}.}.

%which is paramount in real worlds applications
\end{abstract}

%%%%%%%%% BODY TEXT
\section{Introduction}

\begin{figure}
    \centering
    %\vspace{18em}
     \begin{subfigure}[Camera graph]{
     \includegraphics[width=0.46\linewidth]{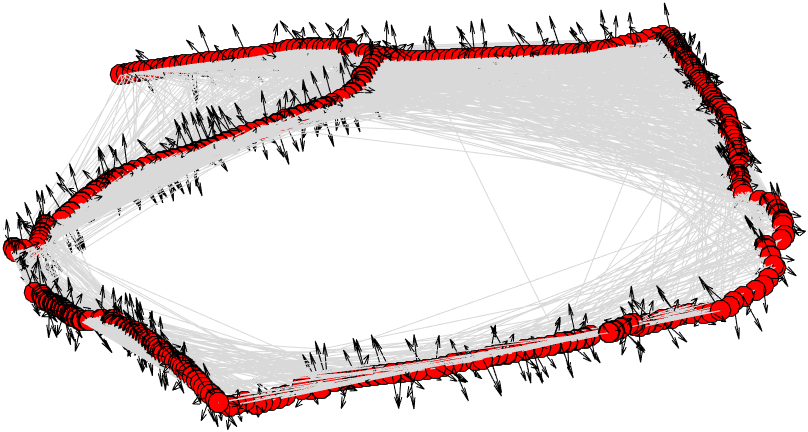}
     \hspace{-0.2cm}
     \label{fig:camera_graph}
     }
     \end{subfigure}
    \begin{subfigure}[Result]{
     \includegraphics[width=0.46\linewidth]{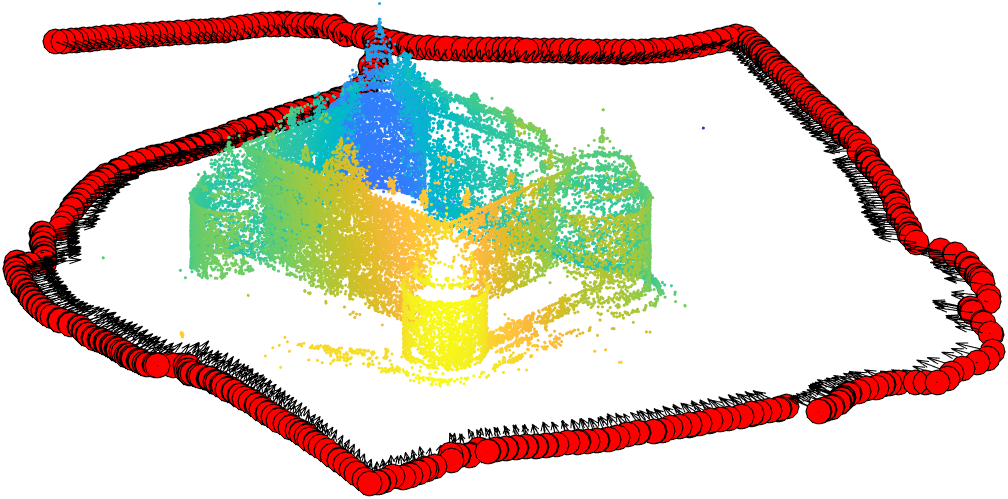}
     %\hspace{6em}
     }
     \end{subfigure}
    
    \caption{(a) Input camera graph from \textit{Orebro Castle}~\cite{olsson11} with $n = 761$ views and $116,589$ connections (relative rotations; grey lines). The initial absolute rotations (represented as black arrows) were randomly chosen. For visualisation, the ground truth positions were used to locate the cameras (red points). (b) Globally optimal absolute rotations computed from our RCD algorithm in $1.96$ s (Shonan averaging~\cite{dellaert20} required $54.62$ s on the same input). Note the alignment of the arrows along the path of the camera (the reconstructed point cloud is also plotted for visualisation).}
    
    % \caption{(a) Input camera graph from \textit{Orebro Castle}~\cite{enqvist2011non} with $n = 761$ views and $116,589$ connections (relative rotations; grey lines). \hlgreen{The relative rotations were chained to produce initial absolute rotations--(random initialisation was used here)}, whose orientations are plotted using black arrows. For visualisation, the ground truth positions were used to locate the cameras (red points). (b) Globally optimal absolute rotations computed from our RCD algorithm in $1.96$ s (the state of the art method of Shonan averaging~\cite{dellaert20} required $54.62$ s on the same input). Note the alignment of the arrows along the path of the camera (the reconstructed point cloud is also plotted for visualisation).}
    
    %For the same instance, Shonan rotation averaging terminates in $30.7$ seconds and BCD in  $XXX$ seconds.}
    \label{fig:sfm}
\end{figure}

Rotation averaging, a.k.a.~multiple rotation averaging~\cite{hartley13} or $SO(3)$ synchronisation~\cite{boumal14}, is the problem of estimating absolute rotations (orientations w.r.t.~a common coordinate system) from a set of relative rotation measurements. In vision and robotics, rotation averaging plays a crucial role in SfM~\cite{martinec07, olsson11, moulon13, cui15, cui17, locher18, zhu18, tron16} and visual SLAM~\cite{carlone15, rosen15, tang17, parra19, li20}, in particular for initialising bundle adjustment. Fig.~\ref{fig:sfm} illustrates the result of rotation averaging. With the increase in the size of SfM problems and continued emphasis on real-time visual SLAM, developing efficient rotation averaging algorithms is an active research area. In particular, real-world applications often give rise to problem instances with thousands of cameras.

%The aim on solving large-scale reconstructions (e.g.~\cite{zhu18} optimises over millions of images) and achieving real-time visual-SLAM systems it makes research on fast rotation averaging algorithms relevant. We present a method that is fast but also globally optimal. 

%\begin{figure}
%    \centering
%    \includegraphics[width=\linewidth]{Fig1.png}
%    \caption{Caption}
%    \label{fig:my_label}
%\end{figure}

%In the context of the latter, 

The input to rotation averaging is a set of noisy relative rotations $\{\tilde{R}_{ij}\}$, where each $\tilde{R}_{ij}$ is a measurement of the orientation difference between cameras $i$ and $j$ which overlap in view. From the relative rotations, rotation averaging aims to recover the absolute rotations $\{R_i\}_{i=1}^n$ which represent the orientations of the cameras. In the ideal case where there is no noise in the relative rotations $\{ R_{ij} \}$,
\begin{align} \label{eq:relrot}
    R_{ij} = R_j  R_i^T.  %R_j = R_{ij} R_i.
\end{align}
The input relative rotations $\{\tilde{R}_{ij}\}$ define a \emph{camera graph} $\cG = (\cV, \cE)$, where $\cV = \{1,\dots,n\}$ is the set of cameras, and $(i,j) \in \cE$ is an edge in $\cG$ if the relative rotation $\tilde{R}_{ij}$ between cameras $i$ and $j$ is measured. We assume a connected undirected graph $\cG$, hence only $\tilde{R}_{ij}$ with $i < j$ needs to be considered. See Fig.~\ref{fig:camera_graph} for an example camera graph.

%Rotation averaging solves for the camera rotations $\{R_i \in SO(3)\}_{i=1}^n$ from measurements $\{\tilde{R}_{ij}\}$ of the relative rotations between pairs of cameras, where the \emph{ideal} relative rotation $R_{ij}$ between two cameras ($R_i$ and $R_j$, $i < j$) comes from the compatibility constraint 
%\begin{align} \label{eq:relrot}
    %R_{ij} = R_j  R_i^T.  %R_j = R_{ij} R_i.
%\end{align}

%The set of relative measurements $\{\tilde{R}_{ij}\}$ forms the so-called \emph{camera graph} $\cG = (\cV, \cE)$ that contains an edge $(i,j) \in \cE$ if there is a relative measurement $\tilde{R}_{ij}$ between the (absolute) rotations $R_i$ and $R_j$. The nodes in the camera graph index the absolute rotations $\cV = \{1,\ldots, n\}$\footnote{We assume that $(\cV, \cE)$ is a connected graph; otherwise, we can solve rotation averaging over multiple independent connected graphs.}. See Fig.~\ref{fig:camera_graph} for a camera graph depiction. 

Rotation averaging is usually posed as a nonlinear optimisation problem with nonconvex domain
\begin{align}\label{eq:convrotavg}
\min_{R_1, \ldots, R_n \in SO(3)} \sum_{(i,j) \in \cE } d(R_j R_i^T, \tilde{R}_{ij})^p,
\end{align}
where $d: SO(3) \times SO(3) \mapsto \mathbb{R}$ is a distance function that measures the deviation from the identity~\eqref{eq:relrot} based on measured and estimated quantities. For example,
\begin{align} \label{eq:chordal}
d_\text{chordal}(R,S) = \| R-S\|_F,
\end{align}
which is known as the chordal distance, and
\begin{align}
d_{\angle}(R,S) = \|\log(RS^T) \|_2
%d_\text{???}(R,S) = ?
\end{align}
which is called the angular distance ($\log : SO(3) \mapsto \mathbb{R}^3$ is the logarithmic map in $SO(3)$~\cite{hartley13}). Also, usually $p=1,2$.

The general form of~\eqref{eq:convrotavg} can be challenging to solve~\cite{hartley13,wilson16}. Earlier efforts devised locally convergent methods~\cite{govindu01, moakher02, govindu04, hartley11, tron12, chatterjee13, hartley13}, e.g., IRLS~\cite{chatterjee13} and the Weiszfeld algorithm~\cite{hartley11}, though most are not able to guarantee local correctness~\cite{wilson16}. In contrast with local methods, convex relaxation methods which include linear method \cite{martinec2007robust}, spectral decomposition methods~\cite{arrigoni16,arrigoni2018robust} and semidefinite method \cite{wang13,singer11} solve a relaxed problem. However, their formulations are either prone to suboptimal solution or the deviation between the relaxed solution and the global solution is unknown.
%solve a relaxed problem optimally, though the deviation between the relaxed solution and the global solution is unknown. 
%Tron \etal~\cite{tron16} surveyed and benchmarked approximate rotation averaging methods in the context of SfM. Recently, learning-based approaches~\cite{purkait20} that can exploit the statistics of camera graphs from an environment have been developed.
Tron \etal~\cite{tron16} surveyed and benchmarked approximate rotation averaging methods in the context of SfM. Recently, learning-based approaches~\cite{purkait20} that can exploit the statistics of camera graphs have been developed.

%There also exist global \hl{A different approach is to 

\subsection{Strong duality}\label{subsec:duality}

Building upon empirical observations (e.g.,~\cite{fredriksson12}),
Eriksson \etal~\cite{eriksson18} proved that the specific version
\begin{align}\label{eq:special}
\min_{R_1, \ldots, R_n \in SO(3)} \sum_{(i,j) \in \cE } d_\text{chordal}(R_j R_i^T, \tilde{R}_{ij})^2,
\end{align}
which is a standard formulation in the literature~\cite{hartley11,hartley13,eriksson18,chatterjee13}, satisfies \emph{strong duality}~\cite{ruszczynski11} under mild conditions on the noise of the input relative rotations (see~\cite[Eq.~(22)]{eriksson18} or the supp. material for details). This means that the global solution to~\eqref{eq:special} can be obtained by solving its Langrangian dual, which is a semidefinite program (SDP) (details in Sec.~\ref{sec:prelim}).

%\paragraph{Solving the SDP relaxation}

Our work focuses on solving the SDP relaxation of~\eqref{eq:special}, especially for large-scale problems. Although SDPs are tractable, generic SDP solvers (e.g., conic optimisation~\cite{sturm99}) can be slow on instances derived from rotation averaging. Thus, exploiting the problem structure to construct faster algorithms is an active research endeavour.

Eriksson \etal~\cite{eriksson18} presented a block coordinate descent (BCD) algorithm to solve the SDP relaxation, which consumed one order of magnitude less time than SeDuMi~\cite{sturm99} on small to moderately sized instances ($n \le 300$). The BCD algorithm maintains and iteratively improves a dense $3n\times 3n$ positive semidefinite (PSD) matrix by updating $3 \times 3n$ submatrices (called ``block rows") until convergence. At convergence, each block row contains rotation matrices (up to correcting for reflection) which are the solution to~\eqref{eq:special} (the solution of different block rows differ by a gauge freedom; see Sec.~\ref{sec:gauge}). However, recent results~\cite{tian19,dellaert20} suggest that BCD is still not practical for large-scale problems encountered in SfM and SLAM, where $n \ge 1000$.

%where $n \ge 200$.

%\hl{However, BCD is not practical for SLAM and large SfM applications}~\cite{tian19}. The scalability issue of BCD radicates in its need to operate and store a dense $3n\times 3n$ matrix. Our approach (Sec.~\ref{sec:method}) does not suffer from this scalability issue as it requires no manipulations of large $3n\times 3n$ square matrices. Instead, our algorithm operates over \hl{``block vectors''}  ($3n\times 3$ matrices). Moreover, our approach updates all rotations at every iteration, which facilitates the incorporation of local methods to speed-up convergence (Sec.~\ref{sec:speedup}). In contrast, BCD updates (``block'') rows of a $3n\times 3n$ matrix until convergence where any row provides the solution\footnote{All rows are equivalent up to some orthogonal matrix.}. However, at intermediate steps, block elements in rows are not orthogonal matrices in general. In contrast, our algorithm keeps updating a block vector of rotations, and facilitates the incorporation of local optimisation methods by directly optimising over the current estimation. 

%that is guaranteed to find the global solution of~\eqref{eq:convrotavg} if the maximum residual fulfill the conditions for strong duality. 

\subsection{Riemannian staircase methods}

%\hl{I will present the line of work of SE-Sync}~\cite{rosen19} (PGO) $\rightarrow$ Tian et al.~\cite{tian19} (PGO) $\rightarrow$ Shonan (specialised for rotation averaging).

The Riemannian staircase framework~\cite{absil09} has been applied successfully to pose graph optimisation (PGO) or $SE(3)$ synchronisation, which aim to recover absolute camera poses (6 DOF) from measurements of relative rigid motion. Under this framework, Rosen \etal.~\cite{rosen19} presented SE-Sync for PGO which guarantees global optimality for moderate noise levels. Tian \etal~\cite{tian19} builds upon SE-Sync to solve PGO in a distributed optimisation setting targetting collaborative SLAM for  multi-robot missions. 

Recently, Dellaert \etal~\cite{dellaert20} adapted SE-Sync for rotation averaging. Their algorithm, called Shonan rotation averaging (henceforth, ``Shonan") can globally solve the SDP relaxation of~\eqref{eq:special}  through a chain of sub-problems on increasingly higher-dimensional domains $SO(d)$, with $d \ge 3$. A certification mechanism checks if the solution of each sub-problem has reached global optimality by computing the minimum eigenvalue of a large $3n \times 3n$ matrix. While optimality is ensured for $d \le 3n+1$, in practice the algorithm only needs to expand $d$ once or twice to reach optimality. Results show that Shonan was an order of magnitude faster than BCD on moderate size instances ($n\leq 200$) and was able to solve large-scale instances ($n \ge 1000$) that were not achievable by BCD with impressive runtimes (instances with $n = 5750$ could be solved in $115$ seconds).

\subsection{Our contributions}

We propose a novel algorithm called \emph{rotation coordinate descent (RCD)} to solve rotation averaging~\eqref{eq:special} globally optimally. Unlike BCD, RCD neither maintains a $3n \times 3n$ dense PSD matrix nor updates the matrix block row-by-block row. Instead, the operation of RCD is equivalent to directly updating the $n$ rotation matrices $R_1,\dots,R_n$, with provable convergence to global optimality. Moreover, since RCD maintains valid rotations at all times, local methods~\cite{chatterjee13,hartley11} can be employed for further speed-ups. 
%see the supplementary material for an implementation of RCD and a demonstration program.

%RCD is also simple to implement; see the supplementary material for C++ implementation of RCD with only $\approx 50$ lines of code and a demonstration program.

We will present results which show that RCD can be \emph{up to two orders of magnitude} faster than Shonan, depending on the structure of the camera graph $\cG$. More specifically, RCD is comparable to Shonan for sparse $\cG$ (\eg, SLAM camera graphs). However, RCD considerably outperforms Shonan on denser graphs (\eg, SfM camera graphs). This makes RCD a much more scalable algorithm.

% We will present results with \hl{ablation tests} which show that RCD can be \emph{up to two orders of magnitude} faster than Shonan, depending on the structure of the camera graph $\cG$. More specifically, RCD is comparable to Shonan for sparse $\cG$ (small number of edges per node, \eg, SLAM camera graphs). However, RCD considerably outperforms Shonan on denser graphs (\eg, SfM camera graphs). This makes RCD a much more scalable algorithm.

\vspace{-1em}
\paragraph{On outliers}

An outlier in rotation averaging~\eqref{eq:convrotavg} is a measured relative rotation $\tilde{R}_{ij}$ that significantly deviates from the true value. Note that formulation~\eqref{eq:special}, i.e., least sum of squared chordal distances, is non-robust. Thus, if there are outliers in the input, BCD, Shonan and RCD will fail, in the sense that they do not return results that closely resemble the ``desired" solutions. In practice, such negative outcomes can be prevented by removing outliers with a preprocessing step~\cite{zach10, enqvist11, olsson11}. We also emphasise that the theoretical validity of our work is not invalidated by the lack of robustness in the standard formulation~\eqref{eq:special}~\cite{hartley11,hartley13,chatterjee13,eriksson18}.

%\hl{we refer the reader to} \cite{yang2020graduated,lajoie2019modeling} for a general outlier rejection approach on SDP problems. 

%which provides global solvers based on SDP in the presence of outliers.

Yang and Carlone~\cite{yang20} proposed a robust SDP relaxation for \emph{single} rotation averaging, a special case where $n=1$ (see~\cite{hartley13}). The method has been demonstrated on relatively small scale problems (less than $100$ measurements). \cite{yang2020graduated, lajoie2019modeling} addressed outliers on SDP based PGO though without providing global optimality guarantees.

%\cite{lajoie2019modeling} proposed a (framework)(graphical model) to incorporate robustness whose resulting formulation is a SDP relaxation  

%\cite{yang20} employed an SDP relaxation over a truncated least-squares (TLS) formulation.

% \hl{Is there any work on global optimality certification in robust rotation averaging by Carlone or others?}
% this one~\cite{yang20} \hl{is related``One Ring to Rule Them All: Certifiably Robust Geometric Perception with Outliers'' as present some application for \emph{single} rotation avg. They target a TLS obj and I think they claim \emph{empirical} optimality... I will read the paper.}

%\hl{Also, I should say something about Wang and Singer}~\cite{wang13} \hl{work: ``Exact and stable recovery of rotations for robust synchronization''.}

%\vfill

%A second strategy is using a robust distance~\cite{chatterjee17}; however, there are not theoretical results for the strong duality over robust formulations. There also exists approximated methods as the low-rank and sparse matrix decomposition formulation in~\cite{arrigoni18}. 

%hus, our method is superior to Shonan over a variety of graph connectivities

%------------------------------------------------------------------------

%\pagebreak

\section{Preliminaries}\label{sec:prelim}

\subsection{Notation}\label{sec:notation}

We operate on block matrices composed of $3 \times 3$ blocks (submatrices in $\mathbb{R}^{3 \times 3}$). A block matrix is represented with a capital letter, e.g., $A \in \mathbb{R}^{3m \times 3n}$, and element $(i,j)$ of a block matrix, denoted $A_{i,j}$, is the submatrix formed by rows $3(i-1) + 1$ to $3(i-1) + 3$ and columns $3(j-1) + 1$ to $3(j-1) + 3$ of $A$. Thus, $A_{i,i}$ are diagonal blocks.

We also define the k-$th$ ``row" of $A$ as the submatrix
\begin{align}
     A_{k,:} =  \left [ A_{k,1} A_{k,2} \cdots A_{k,n} \right] \in \mathbb{R}^{3 \times 3n}
\end{align}
and similarly for the $k$-th ``column" of $A$. If $A$ has a single block column, we call it a ``vector". We use the notation
\begin{align}
    A_{(a:b);(c:d)} = \begin{bmatrix}
    A_{a,c} &\cdots& A_{a,d}\\
    \vdots & & \vdots\\
    A_{b,c} &\cdots& A_{b,d}\\
    \end{bmatrix} \in \mathbb{R}^{3m \times 3n}
\end{align}
for the submatrix of $A$ from rows $a$ to $b$ and columns $c$ to $d$. If $A$ is a vector we use the notation $A_k = A_{k,1}$ and $A_{a:b} = A_{(a:1);(b:1)}$.

% \begin{align}
%     %A_{a:b} = \left[ A_a^T\, A_{a+1}^T\, \cdots  A_{b}^T \right]^T
%     A_{a:b} = A_{(a:1);(b:1)}
% \end{align}
% defines its sub-vector with elements from $a$ to $b$ in $A$.
% %Similarly if $A$ is a column vector.

We denote the $3 \times 3$ identity and zero matrices as $I_3$ and $0_3$, and the trace and Moore–Penrose pseudoinverse of a matrix $M$ as $\tr(M)$ and $M^{\dagger}$, respectively.

%The trace of a square matrix $M$ is denoted as $\tr(M)$. The Moore–Penrose pseudoinverse of $M$ is denoted as $M^{\dagger}$.

\subsection{SDP relaxation}\label{sec:sdp}

We first present the SDP relaxation of~\eqref{eq:special} following~\cite{eriksson18}. By rewriting the chordal distance using trace, \eqref{eq:special} becomes
\begin{align}\label{eq:rotavg}
 	 \min_{R_1, \ldots, R_n \in SO(3)}  -\sum_{(i,j)\in \cE}\tr(R_j^T \tilde{R}_{ij} R_i).
\end{align}
This can be further written more compactly as
\begin{align}\label{eq:p}
\tag{P}
\min_{R\in SO(3)^n} - \tr(R^T \tilde{R} R) 
\end{align}
using matrix notations, where
\begin{align}\label{eq:abs}
%R = [R_1\, R_2 \cdots R_n] \in SO(3)^n,
R = \left[R_1^T\, R_2^T \cdots R_n^T\right]^T \in SO(3)^n
\end{align}
contains the target variables, and $\tilde{R}$ is the $3n\times 3n$ block symmetric matrix with upper-triangle elements $(i,j)$ equal to $\tilde{R}_{ij}^T$ if $(i,j) \in \cE$ and $0_3$ otherwise (diagonal elements are $0_3$'s). Problem~\eqref{eq:p} is called the primal problem.

%The primal problem~\eqref{eq:p} is equivalent to Eq. (11) in~\cite{eriksson18} if using the camera relative rotations definition from~\eqref{eq:relrot}.

%\paragraph{The dual of the dual problem} 
%Eriksson \etal~\cite{eriksson18} designed BCD over the dual of the dual of~\eqref{eq:p}

As derived in Eriksson \etal~\cite{eriksson18}, the dual of the Lagrangian dual of~\eqref{eq:p} is the SDP relaxation
\begin{subequations}
\begin{align}\label{eq:dd}
\tag{DD}
 & \min_{Y \in \mathbb{R}^{3n \times 3n}}&&  - \tr(\tilde{R}Y)\\
& \text{s.t.} &&  Y_{i,i} = I_3,\; i=1,\ldots,n. \label{eq:dd_diag}\\
&&& Y \succeq 0, \label{eq:dd:sdp}
\end{align}
\end{subequations}
where $Y$ is a $3n \times 3n$ PSD matrix, and $Y_{i,i}$ is the $i$-th diagonal block of $Y$. The interested reader is referred to Eriksson \etal for the detailed derivations. It is proven that, under mild conditions (see supp. material), that
\begin{align}\label{eq:sd}
-tr(\tilde{R}Y^*) = -\tr({R^*}^T \tilde{R} R^*),
\end{align}
where $R^*$ and $Y^*$ are respectively the optimisers of~\eqref{eq:p} and~\eqref{eq:dd}, i.e., zero duality gap between~\eqref{eq:p} and~\eqref{eq:dd}.

\begin{algorithm}[t]
\begin{algorithmic}[1]
\REQUIRE $\tilde{R}$ and $Y^{(0)} \succeq 0$.
\STATE $t \leftarrow 0$.
\REPEAT 
\STATE Select an integer $k$ in the interval $[1,n]$. \label{alg:bcd:k} %$k \leftarrow \text{mod}(t,n)$.
\STATE $W \leftarrow $the $k$-th column of $\tilde{R}$. \label{alg:bcd:w}
\STATE $Z \leftarrow  Y^{(t)} W$.\label{alg:bcd:z}
%\STATE $(Z)_k \leftarrow  0_3$.\label{alg:bcd:zero}
\STATE $S \leftarrow Z \left[ \left( W^T Z  \right)^\frac{1}{2} \right]^\dagger $.\label{alg:bcd:s}
% \STATE $Y^{(t+1)} \leftarrow \begin{bmatrix} 
% I_3 & S^T\\
% S & Y^{(t)}
% \end{bmatrix},$ \hl{(succeeded by the appropriate reordering).} \label{alg:bcd:update_a}
\STATE $Y^{(t+1)} \leftarrow \begin{bmatrix} 
Y_{(1:k-1);(1:k-1)}^{(t)} & S_{1:(k-1)} & Y_{(1:k-1);(k+1:n)}^{(t)} \\
S_{1:(k-1)}^T & I_3 & S_{(k+1):n}^T \\
Y_{(k+1:n);(1:k-1)}^{(t)} & S_{(k+1):n} & Y_{(k+1:n);(k+1:n)}^{(t)} \\
\end{bmatrix}$  \label{alg:bcd:update}

% \STATE $R^{(t+1)} \leftarrow \begin{bmatrix}
\STATE $t \leftarrow t+1$.
\UNTIL{convergence}
\RETURN $Y^* = Y^{(t)}$.
\end{algorithmic}
\caption{Block coordinate descent (BCD) for~\eqref{eq:dd}.}
\label{alg:bcd}
\end{algorithm}

\vspace{-1em}
\paragraph{Output rotations}

Note that constraint~\eqref{eq:dd_diag} in~\eqref{eq:dd} merely enforces orthogonality in each diagonal block. Hence, in general a feasible $Y$ for~\eqref{eq:dd} is \emph{not} factorisable as the product of two rotation matrices $R R^T$. It can be shown, however, that the optimiser $Y^*$ of~\eqref{eq:dd} is rank-3~\cite{eriksson18}, which admits the factorisation
\begin{align}\label{eq:yopt_fact}
Y^* = Q^* {Q^*}^T,
\end{align}
where $Q^* \in O(3)^n$ contains $n$ $3\times 3$ orthogonal matrices. To obtain $R^*$, first $Q^*$ is obtained via SVD on $Y^*$, then for each $Q^*_i$ whose determinant is negative, the sign of the $Q^*_i$ is flipped to positive to yield a valid rotation.

% Note that constraint~\eqref{eq:dd_diag} in~\eqref{eq:dd} merely enforces orthogonality in each diagonal block. Hence, in general a feasible $Y$ for~\eqref{eq:dd} is \emph{not} factorisable as the product of two rotation matrices $R^{(t)}R^{(t)}^T$. It can be shown, however, that the optimiser $Y^*$ of~\eqref{eq:dd} is rank-3~\cite{eriksson18}, which admits the factorisation
% \begin{align}\label{eq:yopt_fact}
% Y^* = Q^* {Q^*}^T = Q^*W {Q^*}^T,
% \end{align}
% where $Q^* \in O(3)^n$ contains $n$ $3\times 3$ orthogonal matrices, \hl{and} $W \in \mathbb{R}^{3n \times 3n}$ is an arbitrary orthogonal transformation. To obtain $R^*$ from $Q^*$, the sign of the determinant of each $Q^*_i$ is flipped to positive as required to produce $R^*_i$.

%------------------------------------------------------------------------

\subsection{Block coordinate descent}\label{sec:gauge}

Algorithm~\ref{alg:bcd} presents BCD~\cite{eriksson18} for~\eqref{eq:dd} using our notation, which also includes a minor improvement to the original. Specifically, instead of working on an auxiliary square matrix obtained by removing the $k$-th row and column from $Y^{(t)}$ (see~\cite[Step~3 of Algorithm~1]{eriksson18}), we directly operate over $Y^{(t)}$ and create a temporary block vector $Z$ (Line~\ref{alg:bcd:z}). Since $Z$ is smaller than the auxiliary square matrix, the efficiency of Line~\ref{alg:bcd:s} which requires operating over $Z$ twice is marginally improved. We emphasise that Algorithm~\ref{alg:bcd} is intrinsically the same as the original (see supp. material for details and validity of the improvement).

The PSD matrix $Y$ can be initialised as an arbitrary PSD matrix. A simple choice is setting $R_i = I_3$ for all $i$ in $R$ and initialising $Y^{(0)} = R R^T$. However, we remind again that the subsequent $Y^{(t)}$ are not factorisable as the product of rotations $R^{(t)} {R^{(t)}}^T$ in general; see Sec~\ref{sec:sdp}.

\vspace{-1em}
\paragraph{Gauge freedom}

%Can the aribtarry $T$ be the basis for the gauge freedom, i.e., the latter is a special case of the former???? \hl{yes}

Note that the factorisation~\eqref{eq:yopt_fact} is up to an arbitrary orthogonal transformation $G\in O(3)$, i.e.,
\begin{align}
Y^* = Q^* {Q^*}^T = (Q^ * G) (Q^ * G)^T.
\end{align}
We say that $G$ represents a ``gauge freedom" in the solution. This leads to another approach to retrieve $R^*$ from $Y^*$, which recognises that the columns (and rows) of $Y^*$ are related by orthogonal transformations as $Y^*$ is rank-$3$ with diagonal elements equal to $I_3$. Thus, for any two columns $k$ and $k'$ in $Y^*$, there exists an orthogonal transformation $G_{k,k'}\in O(3)$ such that
\begin{align}
    Y^*_{:,k'} =  Y^*_{:,k} \; G_{k,k'} .
\end{align}
Hence, $G_{k,k'}$ must transform the $k'$-th element of $Y^*_{:,k}$ to $I_3$ (i.e., $Y^*_{k',k} G_{k,k'} = I_3$). Therefore
\begin{align}\label{eq:tsp}
    G_{k,k'} = (Y^*_{k',k})^T = Y^*_{k,k'}
\end{align}
as columns in $Y^*$ are orthogonal and $Y^*$ is symmetric.

The set of transformations relating columns~\eqref{eq:tsp}
\begin{align}
    \mathscr{G} = \left\{ G_{k,k'},\,   \text{ for all }  k,k'=1,\ldots,n \right\} \subset O(3)
\end{align}
corresponds to an special case of gauge freedom. Since all columns in $Y^*$ are up to some transformation in $\mathscr{G}$ to another column, we can take any as $R^*$; the choice will depend on selecting one of the cameras as the reference frame, i.e., which camera takes $R^*_i=I_3$.

\section{Rotation coordinate descent}\label{sec:method}

%BCD requires to maintain a square and dense matrix $Y$ which is not scalable.

%Note also that BCD needs to store the large $\tilde{R}$ matrix that encodes the relative measurements. However, $\tilde{R}$ is sparse for most real-life applications, which points to redundant 
In this section, we will describe our novel method  called \emph{rotation coordinate descent (RCD)}, summarised in Algorithm~\ref{alg:rcd}. While seemingly a minor modification to BCD, RCD is based on nontrivial insights (Sec.~\ref{sec:main_idea}). More importantly, a major contribution is to mathematically prove the global convergence of RCD (Sec.~\ref{sec:global}). Another fundamental advantage is that since RCD maintains valid rotations throughout the iterations (in contrast to BCD; see Sec.~\ref{sec:gauge}), it can exploit local optimisation routines for~\eqref{eq:p} to speed-up convergence (Sec.~\ref{sec:speedup}). As the results will show (Sec.~\ref{sec:experiments}), our approach can be up to two orders of magnitude faster than Shonan~\cite{dellaert20}, which is the state of the art for~\eqref{eq:dd}.

\subsection{Main ideas}\label{sec:main_idea}

\begin{algorithm}[t]
\begin{algorithmic}[1]
\REQUIRE $\tilde{R}$ and $R^{(0)}$.
\STATE $t \leftarrow 0$.
\REPEAT 
\STATE Select an integer $k$ in the interval $[1,n]$. \label{alg:rcd:k}
\STATE $W \leftarrow $the $k$-th column of $\tilde{R}$. \label{alg:rcd:w}
\STATE $Z \leftarrow R^{(t)} ({R^{(t)}}^T W)$.\label{alg:rcd:z}
\STATE $S \leftarrow Z \left[ \left( W^T Z  \right)^\frac{1}{2} \right]^\dagger $.\label{alg:rcd:s}
\STATE $Q^{(t+1)} \leftarrow \left[(S_{1:(k-1)})^T \;\; I_3 \;\; (S_{(k+1):n})^T \right]^T$.\label{alg:rcd:update}
%\STATE $R^{(t+1)} \leftarrow \left[S_1^T \;\; S_{k-1}^T I_3 \;\; (S_{(k+1):n})^T \right]^T$.\label{alg:rcd:update}
%\STATE $R^{(t+1)} \leftarrow$ \hl{Flip determknants.}
\STATE $R^{(t+1)} \leftarrow$ Flip determinants over $Q^{(t+1)}$ (if needed) to ensure rotations.
\STATE $t \leftarrow t+1$. \label{alg:rcd:flip}
\UNTIL{convergence}
\RETURN $Y^* = R^{(t)}{R^{(t)}}^T$.
\end{algorithmic}
\caption{Rotation coordinate descent (RCD) for~\eqref{eq:dd}.}
\label{alg:rcd}
\end{algorithm}

%Say something like the ``effective variables" is just $3n$.... Without the need of operating over $Y$, BCD could have been faster. 

%($Y^* = R^* {R^*}^T$)
% for~\eqref{eq:p}
As summarised in Algorithm~\ref{alg:bcd}, BCD requires to maintain and operate on a large dense PSD matrix $Y \in \mathbb{R}^{3n \times 3n}$. While the values of each update can be computed in constant time (specifically, SVD of a $3\times 3$ matrix; Line~\ref{alg:bcd:s}), manipulating $Y$ is unwieldy. Specifically, Line~\ref{alg:bcd:z} performs
\begin{align}
    Z = Y^{(t)}W
\end{align}
to obtain temporary vector $Z \in \mathbb{R}^{3n \times 3}$ from a subset of the measurements $W \in \mathbb{R}^{3n \times 3}$, which costs
\begin{align}
    27n^2~\text{multiplications} \equiv \mathcal{O}(n^2).
\end{align}
This quadratic dependence on $n$ makes BCD slow on large-scale SfM or SLAM problems~\cite{tian19}, e.g., where $n \ge 1000$, as will be demonstrated in Sec.~\ref{sec:experiments}.

Although the PSD matrix $Y$ of~\eqref{eq:dd} has size $3n \times 3n$, the ``effective" variables are only $3n$ given that $Y^*$ is rank-$3$. Our key insight comes from the  gauge freedom of  $Y^*$ (Sec.~\ref{sec:gauge}) implying that any row of $Y^*$ provides a valid solution for $R^*$.  Choosing the $k$-th row implies choosing the $k$-th camera as the reference frame, i.e., $R_k = I_3$. Based on this insight, we devised RCD to maintain only the effective variables $R^{(t)}$. Each iteration  executes what amounts to updating a single column of $Y$; specifically, in Line~\ref{alg:rcd:k}, a camera $k$ is chosen as the reference frame then set the $k$-th element of $Q^{(t+1)}$ as $I_3$ in Line~\ref{alg:rcd:update} ($Q^{(t)}$ contains orthogonal matrices). Then, in Line~\ref{alg:rcd:s} the other elements of $Q^{(t+1)}$ are updated via the same explicit form of BCD. To ensure keeping rotations elements during iterations, the sign of the orthogonal elements in $Q^{(t+1)}$ is flipped if negative in Line~\ref{alg:rcd:flip} to produce $R^{(t+1)}$.

Maintaining and updating only $R^{(t)}$ provides immediate computational savings; in Line~\eqref{alg:rcd:z} obtaining the intermediate vector $Z$ is now accomplished as
\begin{align}
    Z = R^{(t)} \underbrace{( {R^{(t)}}^T W)}_{\text{Compute this first}},
\end{align}
which costs
\begin{align}
    27n + 27n~\text{multiplications} \equiv \mathcal{O}(n)
\end{align}
and has only linear dependence on $n$. The next section proves the important result that this computational savings does not come at the expense of global optimality.

\subsection{Global convergence of RCD}\label{sec:global}

As proven in~\cite{eriksson18, eriksson19}, Algorithm~\ref{alg:bcd} monotonically decreases the objective $-\tr(\tilde{R}Y)$ at each iteration from any feasible initialisation. Our strategy for proving the global convergence of RCD is to show that updating the variables at each iteration $t$ of Algorithm~\ref{alg:rcd}, i.e.,
\begin{align}
R^{(t)} \rightarrow R^{(t+1)},
\end{align}
has an effect on $-\tr(\tilde{R}Y)$ that is equivalent to one iteration of Algorithm~\ref{alg:bcd} initialised with
\begin{align}\label{eq:y0}
    Y^{(0)} = R^{(t)} {R^{(t)}}^T.
\end{align}
If this equivalence can be established, Algorithm~\ref{alg:rcd} also provably monotonically decreases $-\tr(\tilde{R}Y)$ and will converge to the optimiser $Y^*$ of~\eqref{eq:dd}.

To this end, we will first show (Corollary~\ref{coro:1}) that one iteration of Algorithm~\ref{alg:bcd} initialised with~\eqref{eq:y0} produces a PSD matrix $Y^{(1)}$ that is factorisable as
\begin{align}\label{eq:y1_fact}
    Y^{(1)} = R^{(1)}{R^{(1)}}^T.
\end{align}

Without loss of generality, we take $k=1$ (the updated row and column in BCD during the iteration) and define $R^{(1)}_{\text{BCD}}$ as the first column of $Y^{(1)}$, i.e.,
\begin{align}
R^{(1)}_{\text{BCD}} = Y^{(1)}_{:,1}.
\end{align}
Then, we will prove that $R^{(t+1)} = R^{(1)}_{\text{BCD}}$ (Theorem~\ref{theo:2}). From Line~\ref{alg:bcd:update} in Algorithm~\ref{alg:bcd}, $Y^{(1)}$ can be written as
\begin{align}\label{eq:y1}
	Y^{(1)} = \begin{bmatrix}
	I_3 &{X^*}^T\\
	X^* &B
	\end{bmatrix},
\end{align}
where $B = Y^{(0)}_{(2:n);(2:n)}$ is the unchanged sub-matrix during the iteration $Y^{(0)} \rightarrow Y^{(1)}$, and $X^* \in \mathbb{R}^{3(n-1) \times 3}$ contains the updated values. From~\cite{eriksson18, eriksson19}, $X^*$ is the optimiser of the following SDP problem:
\begin{subequations}\label{eq:s}
	\begin{align}
	 &\min_{X\in\mathbb{R}^{3(n-1)\times 3}} && - \tr(C^T X) &\\
	 &\text{s.t.} && \begin{bmatrix}
	I_3 & X^T\\
	X & B
	\end{bmatrix} \succeq 0, \label{eq:subprob:const}
	\end{align}
\end{subequations}
where $C \in \mathbb{R}^{3(n-1)\times 3 } $ is equal to $W$ as in Line~\ref{alg:bcd:w} in Algorithm~\ref{alg:bcd} but without the $k$-th element (which is zero).

Note that the optimal PSD matrix in Problem~\eqref{eq:s} is $Y^{(1)}$ \eqref{eq:y1}. The goal of Problem~\eqref{eq:s} is to find the optimal update $X^*$ to produce $Y^{(1)}$ that remains feasible (constraint~\eqref{eq:subprob:const}).

\begin{theorem} \label{theo:1}
Problem~\eqref{eq:s} is a special case of~\eqref{eq:dd}.
\end{theorem}

\begin{proof}
Consider the instance of Problem~\eqref{eq:dd} with 
\begin{align}\label{eq:r_sp}
\tilde{R} = \begin{bmatrix}
0_3 & C^T\\
C & 0
\end{bmatrix}. %:= \tilde{R}_k.
\end{align}
We first show that a feasible PSD matrix in Problem~\eqref{eq:s}
\begin{align}\label{eq:y}
Y = \begin{bmatrix}
I_3 & X^T\\
X & B
\end{bmatrix} %:= Y_k
\end{align}
is a feasible solution in~\eqref{eq:dd}. From the initialisation of $Y^{(0)}$ in~\eqref{eq:y0},
%\begin{align}\label{eq:b}
$B = R_{2:n}^{(t)} \, {R_{2:n}^{(t)}}^T$ ;
%\end{align}
hence, all diagonal elements in $Y$~\eqref{eq:y} are identities which fulfill the first  constraint~\eqref{eq:dd_diag} in \eqref{eq:dd}. From~\eqref{eq:subprob:const}, $Y \succeq 0$, which is the second constraint~\eqref{eq:dd:sdp} in \eqref{eq:dd}. 

We now show the objective of~\eqref{eq:dd} with $\tilde{R}$ from~\eqref{eq:r_sp} is equivalent to the objective in Problem~\eqref{eq:s}. The objective of ~\eqref{eq:dd} becomes
\vspace{-2mm}
\begin{subequations}
\begin{align}
    &-\tr(\tilde{R} Y) &=& - \tr\left( \begin{bmatrix}
    C^T X  &  C^T B \\
    C  &  C X^T 
    \end{bmatrix} \right) \\
    &&=& -\tr\left( C^T X \right) -\tr\left( C X^T \right) \\
    &&=& -2 \tr\left( C^T X \right) 
\end{align}
\end{subequations}
which is twice to the objective of~\eqref{eq:s}. Thus, Problem~\eqref{eq:s} is a special case of~\eqref{eq:dd} since any feasible solution of~\eqref{eq:s} is also feasible in~\eqref{eq:dd}, and  both objectives are equivalent. 
\end{proof}

\vspace{-4.5mm}
\begin{lemma} \label{lem:1}
The optimal PSD matrix of Problem~\eqref{eq:s} admits the factorisation
\vspace{-2mm}
\begin{align} \label{eq:fact}
    Y^{*(1)} = R^{*(1)} {R^{*(1)}}^T.
\end{align}
\end{lemma}
\begin{proof}
Problem~\eqref{eq:s} is a special case of~\eqref{eq:dd} (Theorem~\ref{theo:1}) $\implies$ the optimal PSD matrix of Problem~\eqref{eq:s} also admits the factorisation~\eqref{eq:fact}.
\end{proof}

\vspace{-3mm}
\begin{corollary}~\label{coro:1}
Lemma~\ref{lem:1} validates~\eqref{eq:y1_fact} as BCD optimally solves Problem~\eqref{eq:s} (Lines~\ref{alg:bcd:z}--\ref{alg:bcd:update} in Algorithm~\ref{alg:bcd})~\cite{eriksson18,eriksson19}. Thus, Algorithm~\ref{alg:rcd} preserves the rank-$3$ factorisation during iterations, and also Algorithm~\ref{alg:bcd} if initialised with a rank-$3$ SDP matrix $R^{(0)} {R^{(0)}}^T$ with $R^{(0)} \in SO(3)^n$.
\end{corollary}

\begin{theorem}\label{theo:2}
$R^{(t+1)} = R^{(1)}_{\text{BCD}} $
\end{theorem}

\begin{proof}
The equality is by construction of Algorithm~\ref{alg:rcd}. From the definition of $R^{(1)}_{\text{BCD}}$ in~\eqref{eq:y0}, $R^{(1)}_{\text{BCD}}$ is the first column in $Y^{(1)}$~\eqref{eq:y1}, i.e., 
%\begin{align}
  	$R_{\text{BCD}}^{(1)} = \left[I_3\;\; {X^*}^T\right]^T$,
%\end{align}
where $X^*$ is $S$ (in Line~\ref{alg:bcd:s}, Algorithm~\ref{alg:bcd}) without the $k$-th element (see sup. material for details). Lines~\ref{alg:bcd:k}--\ref{alg:bcd:s} in Algorithm~\ref{alg:bcd} are the same as Lines~\ref{alg:rcd:k}--\ref{alg:rcd:s} in Algorithm~\ref{alg:rcd} except on obtaining $Z$, which takes the same value since from the initialisation~\eqref{eq:y0} of $Y^{(0)}$ in Algorithm~\ref{alg:bcd}, 
%\begin{align}
   $ Z =  Y^{(0)} W = R^{(t)} ({R^{(t)}}^T W)$
%\end{align}
is equal to $Z$ as obtained in Algorithm~\ref{alg:rcd} $\implies$ also
%\begin{align}
    $R^{(t+1)} = \left[I_3\;\; {X^*}^T\right]^T$.
%\end{align}
\end{proof}
\section{Speeding up RCD with local optimisation}~\label{sec:speedup}
%\section{Speeding up Algorithm~\ref{alg:rcd} with local optimisation}~\label{sec:speedup}
Since Algorithm~\ref{alg:rcd} iterates over $SO(3)^n$, local methods for~\eqref{eq:p} can be directly used to speedup convergence of Algorithm~\ref{alg:rcd}. Contrast this to BCD that updates a PSD matrix from where, in general, valid rotations can be retrieved only at convergence. Algorithm~\ref{alg:fcd} integrates a local method (Line~\ref{alg:fcd:local}) that we design from experimental observations:
\begin{enumerate} [topsep=0.25em,itemsep=0.25em,parsep=0em,leftmargin=1em]
    \item Substantial reductions in the objective often occur after $n$ iterations. We call it an \emph{epoch} and we ensure we sample all $k$'s during each epoch (Line~\ref{alg:fcd:k}).
    
    \item In practice, one iteration of Algorithm~\ref{alg:rcd} takes $\approx 0.02\%$ of the runtime of solving a local optimisation instance. Thus, Algorithm~\ref{alg:fcd} invokes local optimisation and check for convergence only after completing epochs.
    
    \item Local optimisation produces more drastic ``jumps'' in the objective at earlier iterations. Thus,  Algorithm~\ref{alg:fcd} delays local optimisation when the local method fails on reducing the objective (Line~\ref{alg:fcd:delay}). 
\end{enumerate}

\begin{algorithm}[t]
\begin{algorithmic}[1]
\REQUIRE $\tilde{R}$ and $R^{(0)}$.
\STATE $t \leftarrow 0$, $e \leftarrow 0$, $s \leftarrow 0 $
\REPEAT 
\FOR {$i=1,\ldots, n$}
\STATE Select an integer $k$ in the interval $[1,n]$ w/o rep. \label{alg:fcd:k}

\STATE $W \leftarrow $the $k$-th column of $\tilde{R}$. \label{alg:fcd:w}
\STATE $Z \leftarrow R^{(t)} ({R^{(t)}}^T W)$.\label{alg:fcd:z}
\STATE $S \leftarrow Z \left[ \left( W^T Z  \right)^\frac{1}{2} \right]^\dagger $.\label{alg:fcd:s}
\STATE $R^{(t+1)} \leftarrow \left[(S_{1:(k-1)})^T \;\; I_3 \;\; (S_{(k+1):n})^T \right]^T$.\label{alg:fcd:update}
\STATE $t \leftarrow t+1$.
\ENDFOR
%\IF{$s>e$}
\IF{$(\,s = 0  \, \text{or}\,\,\text{MOD}(e,s) = 0 )$}
\STATE $R^{(t)} \leftarrow$ Flip determinants over $R^{(t)}$ (if needed) to ensure rotations.
\STATE $\hat{R} \leftarrow $ local method with initial estimate $R^{(t)}$.
%\STATE $u \leftarrow 0$.
\label{alg:fcd:local}
\IF{ $-\tr(\hat{R}^T \tilde{R} \hat{R}) < -\tr({R^{(t)}}^T \tilde{R} R^{(t)})$ }
\STATE $R^{(t)} \leftarrow \hat{R}$.
\ELSE
%\STATE $s \leftarrow 2s$. 
\STATE $s \leftarrow s+2$  \label{alg:fcd:delay}
\ENDIF
\ENDIF
\STATE $e \leftarrow e+1$.
%\STATE $u \leftarrow u+1$.
\UNTIL{convergence}
\end{algorithmic}
\caption{RCD with local optimisation (RCDL).}
\label{alg:fcd}
\end{algorithm}

\begin{figure}
    \centering
    \vspace{-2.5mm}
    \begin{subfigure}[]{\includegraphics[width=0.60\linewidth]{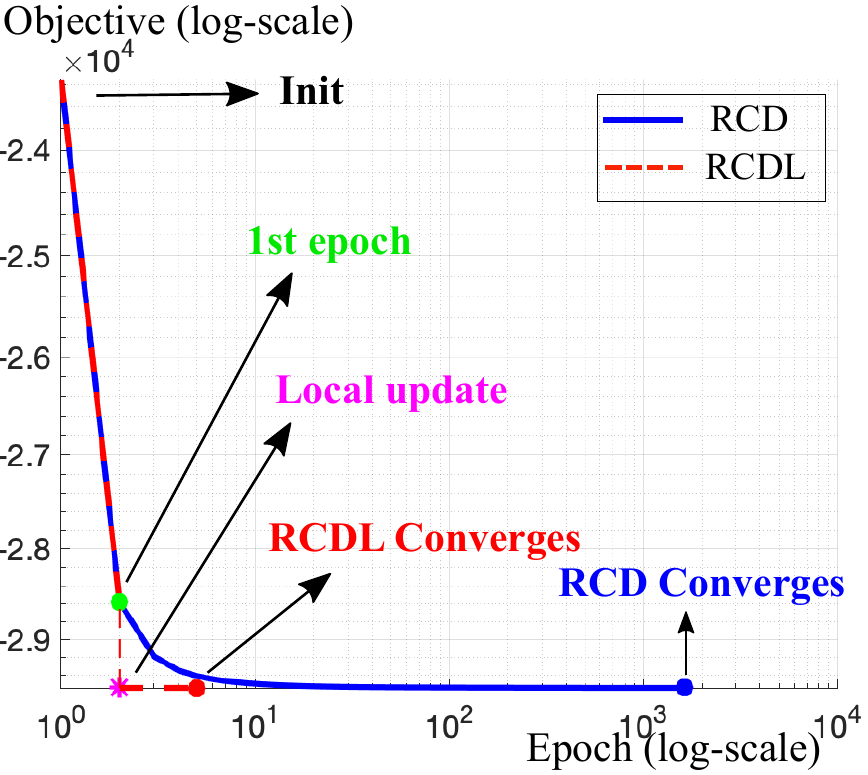}}\end{subfigure}
    \begin{subfigure}[]{\includegraphics[width=0.35\linewidth]{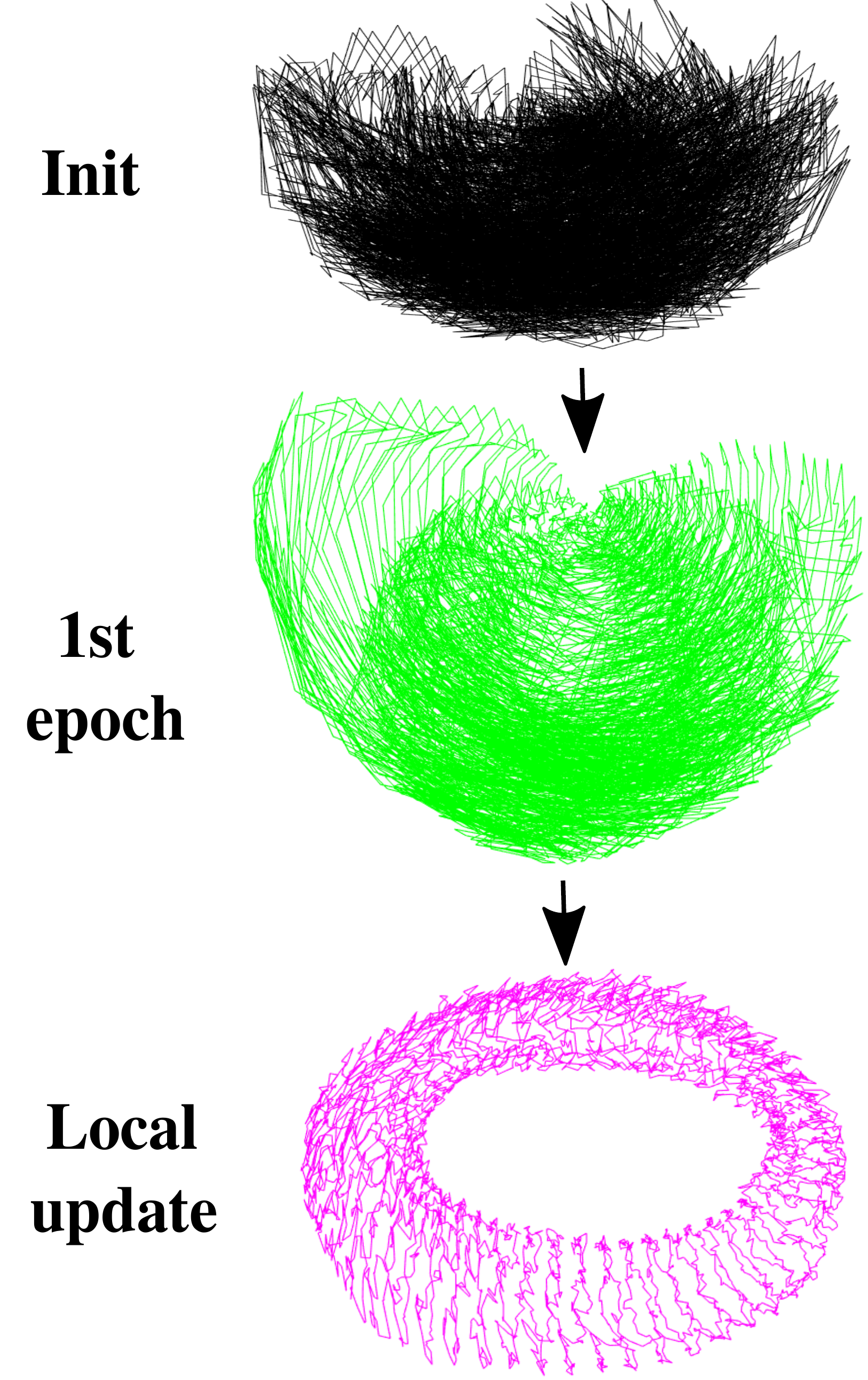}}\end{subfigure}
    \caption{Evolution of RCD and RCDL on the large-scale SLAM instance \emph{torus}~\cite{carlone15} with $n=5000$ cameras. (a) Evolution of the objectives. (b) Camera poses from RCDL. A \emph{single local update} was able to produce a visually correct solution. }
    
    % \caption{\hl{Evolution} of RCD and RCDL on \emph{torus} (details in Table~\ref{table:Slam_benchmark_result}). (a) Evolution of the objectives. (b) Camera poses from RCDL. A \emph{single local update} produced a visually correct result. }
    \label{fig:slam_RCD_vs_RCDL}
\end{figure}

To demonstrate the effect of local optimisation on the convergence of RCD, Fig.~\ref{fig:slam_RCD_vs_RCDL} plots the objective value for RCD and RCDL at increasing epochs on the input graph \emph{torus}~\cite{carlone15} with $n = 5000$ cameras (see Table~\ref{table:Slam_benchmark_result} in Sec.~\ref{sec:experiments} for more details). During the 1st epoch, the local algorithm drastically reduced the objective (from stage in \emph{green} to stage in \emph{magenta}). This ``jump'' of the objective value reveals the collaborative strength of global and local methods, which enabled RCDL~to converge in much fewer epochs (\emph{red} stage) compared to RCD (\emph{blue} stage).

% \hl{To demonstrate} the effect of local optimisation on the convergence of RCDL, Fig.~\ref{fig:slam_RCD_vs_RCDL} plots the objective value for RCD and RCDL on \emph{torus} at increasing epochs. During the 1st epoch, the local algorithm drastically reduces the objective (from point in \textit{green} to point in \textit{magenta}). This reveals the collaborative strength of global and local methods, which enables RCDL~\hlgreen{to} converge in much fewer epochs (\textit{red dot}) compared to RCD (\textit{blue dot}). 

%We remark that the convergence of RCDL could be improved by utilising more efficient local iterative solvers.

% RCD vs RCDL

%Since Algorithm~\ref{alg:rcd} iterates over elements of $SO(3)^n$ (if choosing signs for the elements of the solution to ensure positive determinants), the use of a local rotation averaging methods to speedup convergence is direct. Contrast to BCD that updates a PSD matrix from which, in general, rotations can be retrieved only at convergence. 

%without no requiring computing an approximate initial estimate of rotations from  before invoking a local method. 

%---------------------------------------------------------------------------------------------
\section{Experiments}\label{sec:experiments}
%We benchmarked the following rotation averaging methods:
% We benchmark the following algorithms:
% \begin{itemize} [topsep=0.25em,itemsep=0.25em,parsep=0em]
%     \item BCD: Algorithm~\ref{alg:bcd}.
%     \item RCD: Algorithm~\ref{alg:rcd}.
%     \item RCDL: Algorithm~\ref{alg:fcd} with local optimisation routine adapted from~\cite{parra19}.
%     \item SA: Shonan rotation averaging~\cite{dellaert20}.
% \end{itemize}

We benchmarked the following algorithms over a variety of synthetic and real-world camera graph inputs: Algorithm~\ref{alg:bcd} (BCD), Algorithm~\ref{alg:rcd} (RCD), Algorithm~\ref{alg:fcd} (RCDL) with local optimisation routine adapted from~\cite{parra19}, and Shonan~\cite{dellaert20} (SA). We implemented all the optimisation routines in C++ except for SA for which we used the author's implementation (which also has optimisation routines in C++ \footnote{\url{https://github.com/dellaert/ShonanAveraging}}). 
As we measured runtime as the total algorithm time, we include the certification times in SA. All experiments are executed on a standard machine with an Intel Core i5 2.3 GHz CPU and 8 GB RAM.

\begin{figure*}%[th]
	\begin{tabularx}{\textwidth}{c|C{.29} C{.29} C{.29}}
		& {\small $n=100$} & {\small $n=200$} & {\small $n=1000$}\\
		
		\hline
		& & & \\[-1em]
		
		\rotatebox[origin=l]{90}{\scriptsize \hspace{1cm} \emph{SfM}}&
		\includegraphics[width=0.29\linewidth, trim=0 0 .15cm 0, clip]{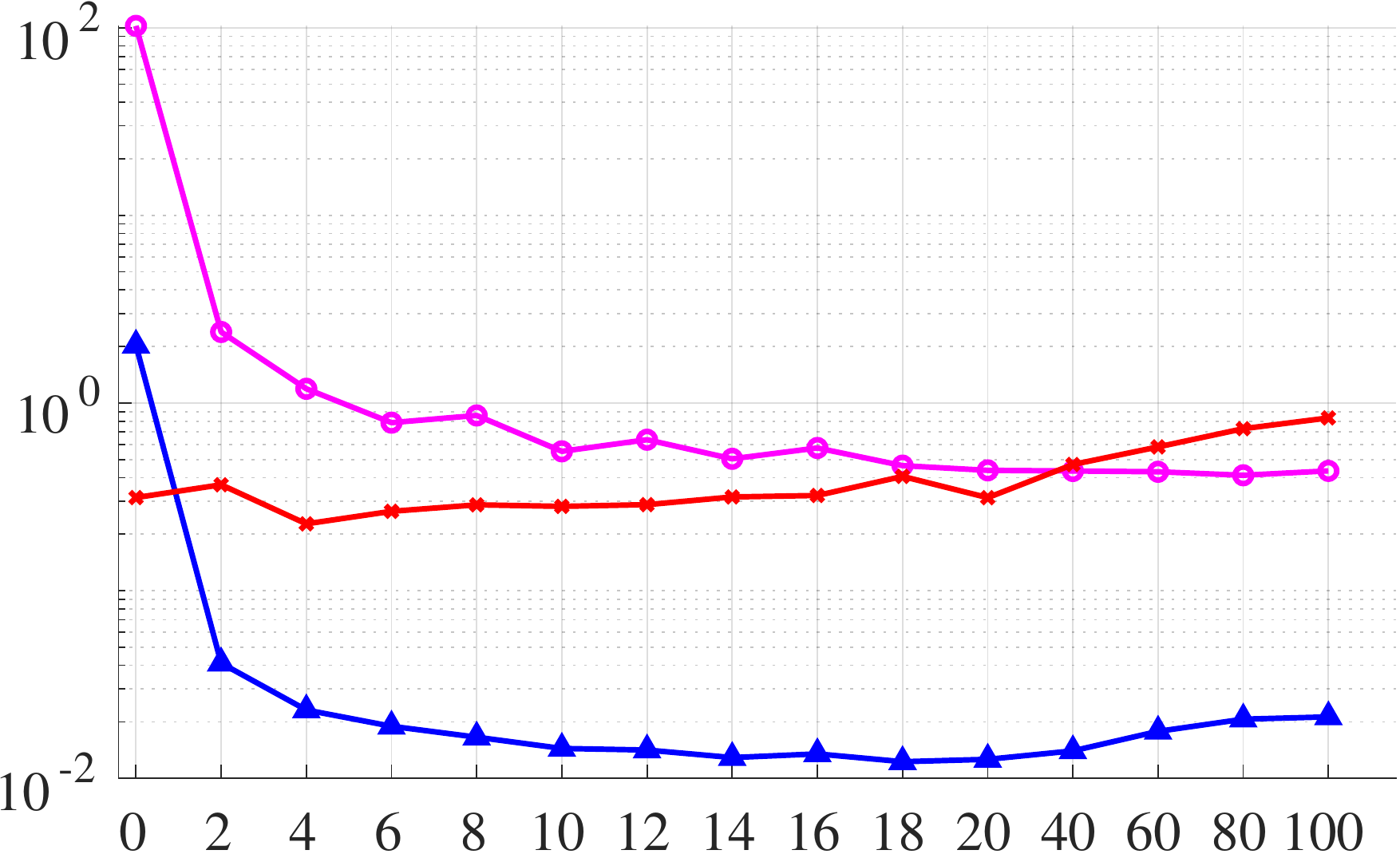} &
		\includegraphics[width=0.29\linewidth, trim=0 0 .15cm 0, clip]{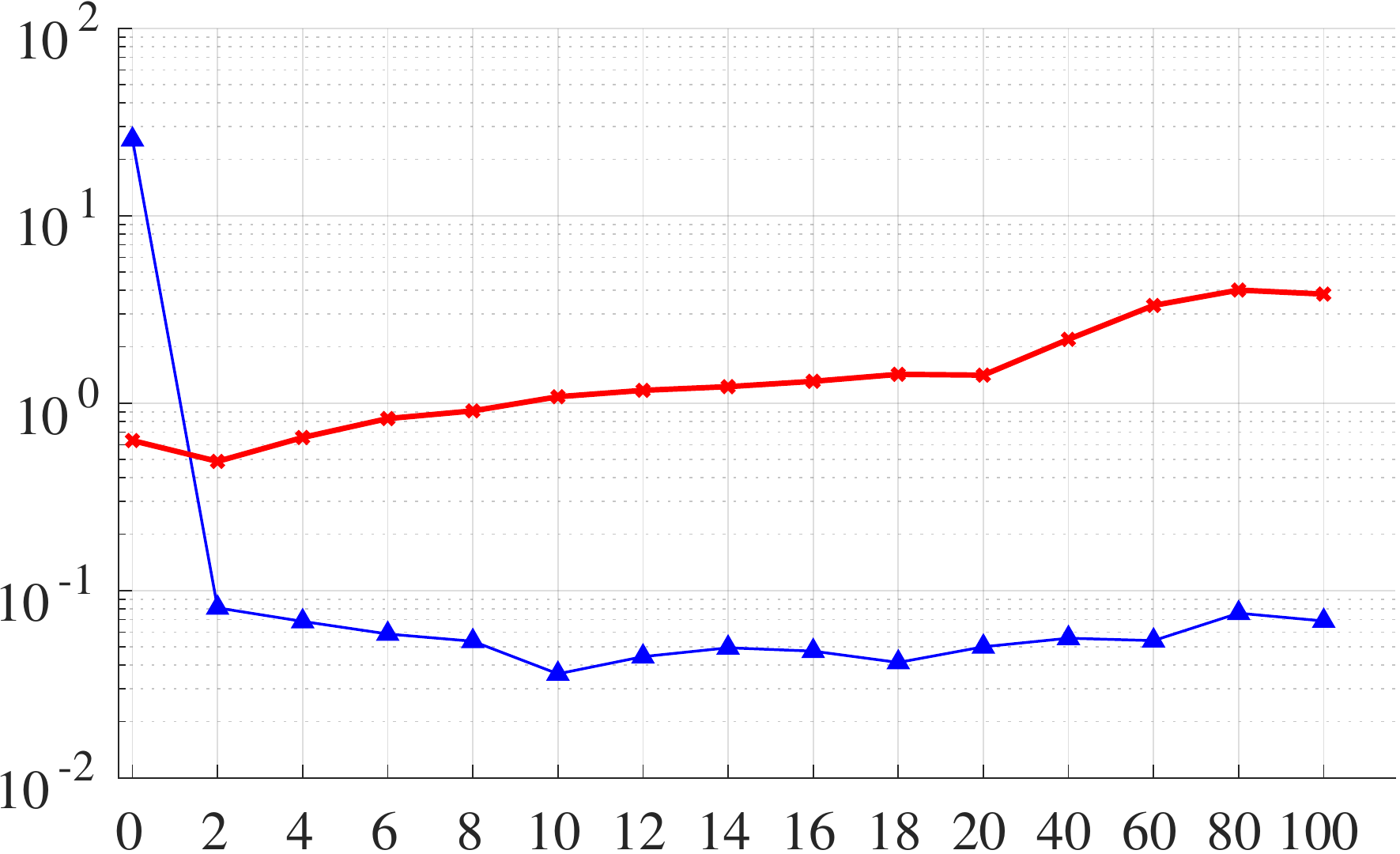} &
		\includegraphics[width=0.29\linewidth, trim=0 0 .15cm 0, clip]{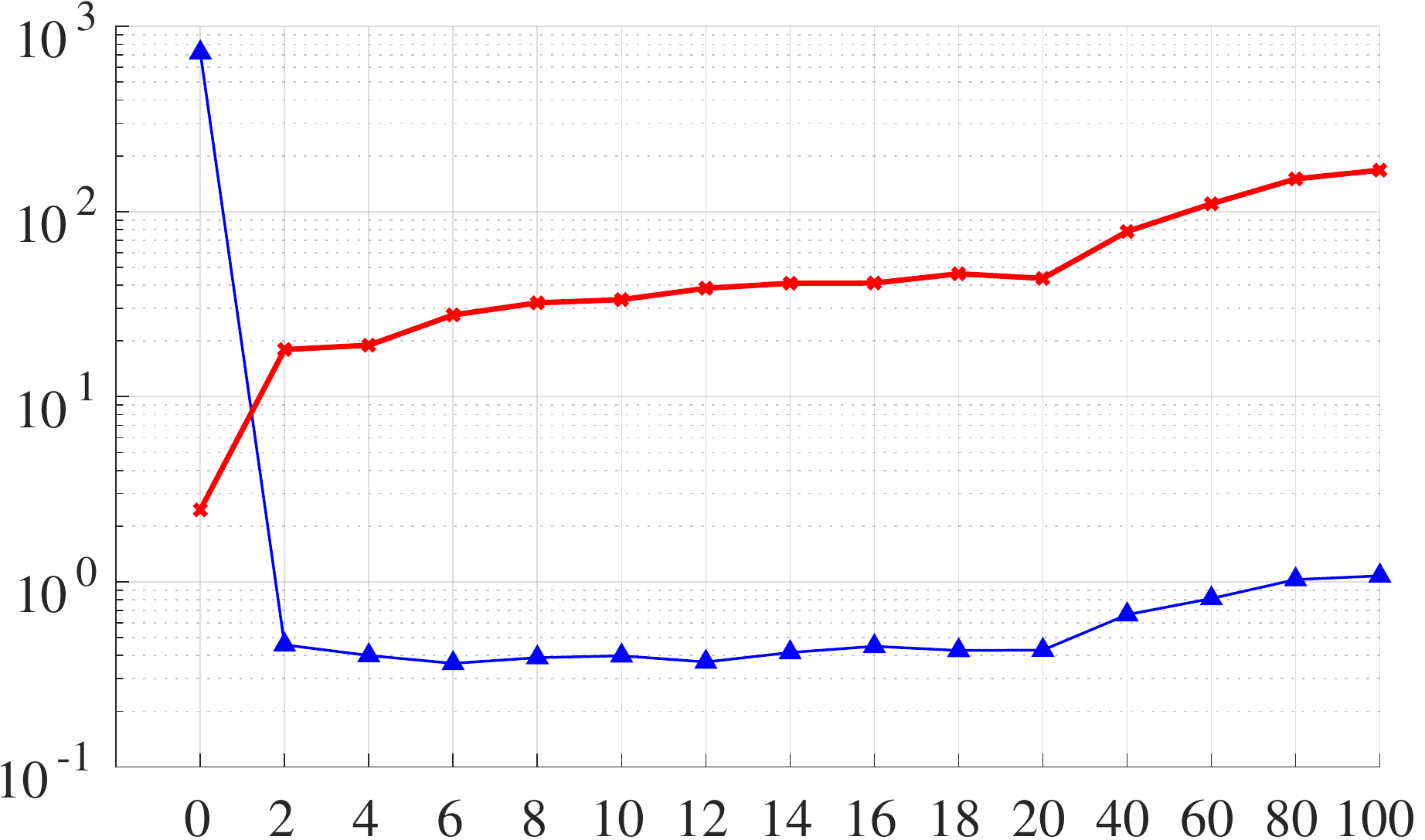} \\
		[1mm]
		%\hline
		
		\rotatebox[origin=l]{90}{\scriptsize \hspace{1cm} \emph{SLAM}}&
		\includegraphics[width=0.29\linewidth, trim=0 0 .15cm 0, clip]{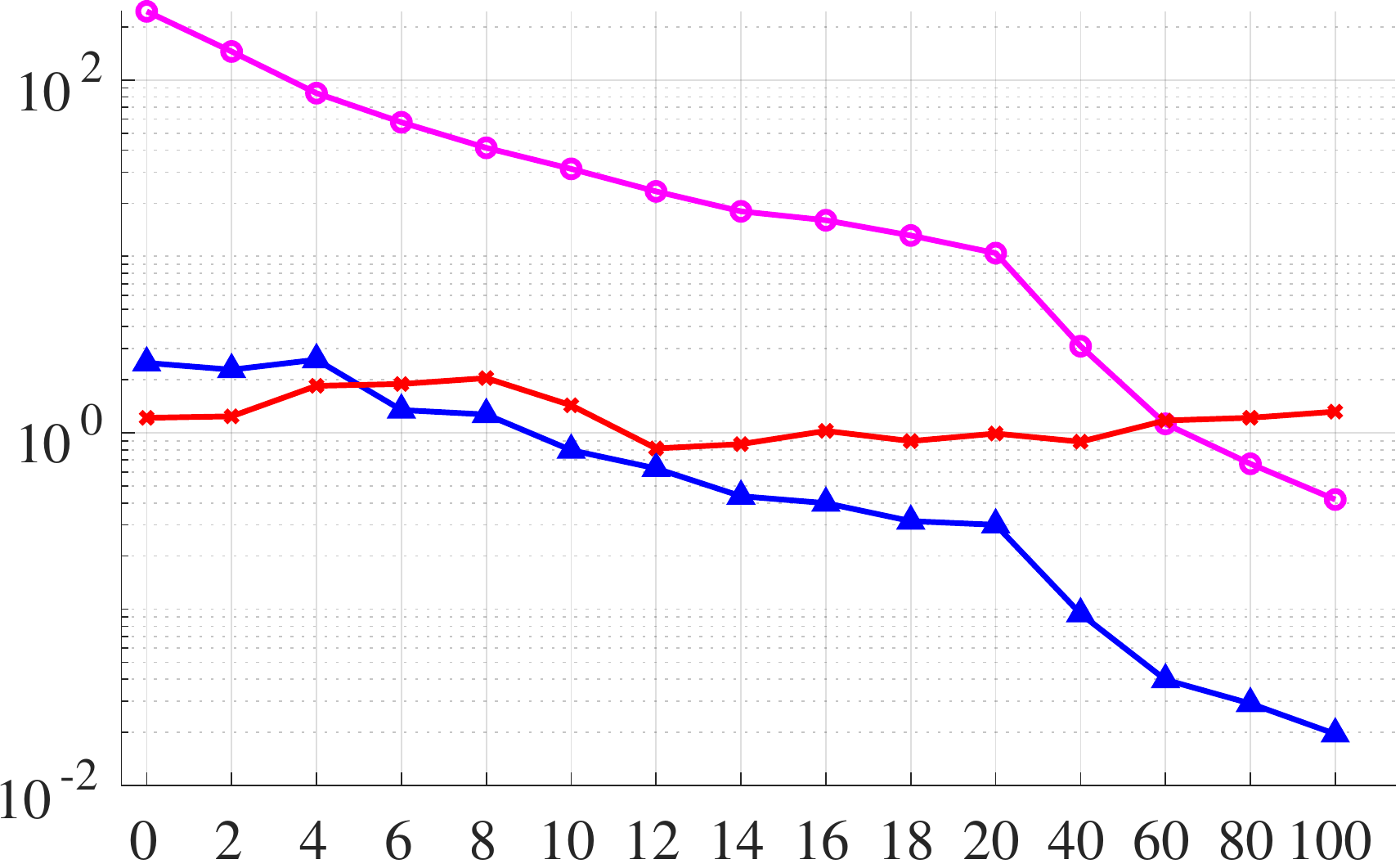} &
		\includegraphics[width=0.29\linewidth, trim=0 0 .15cm 0, clip]{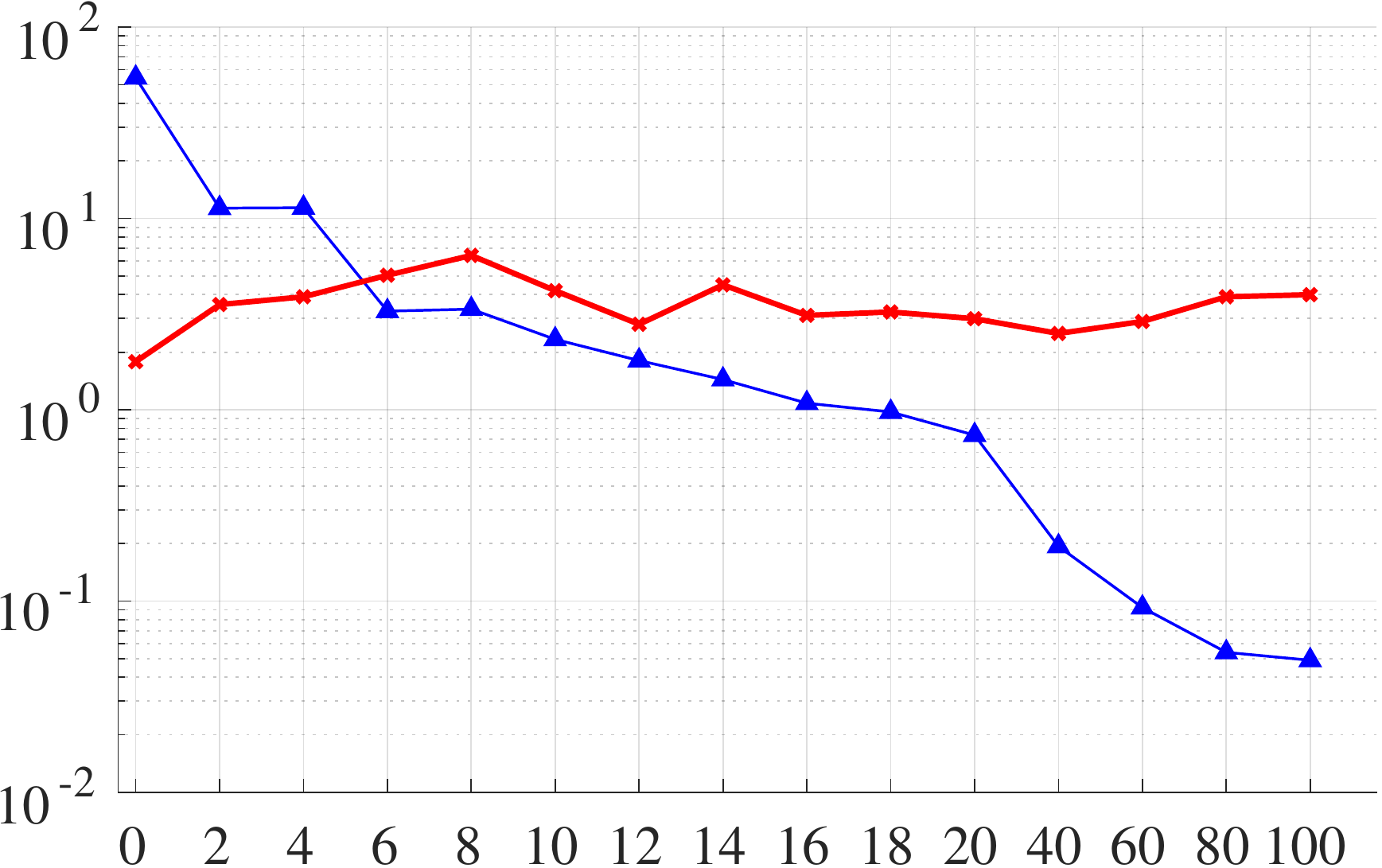} &
		\includegraphics[width=0.29\linewidth, trim=0 0 .15cm 0, clip]{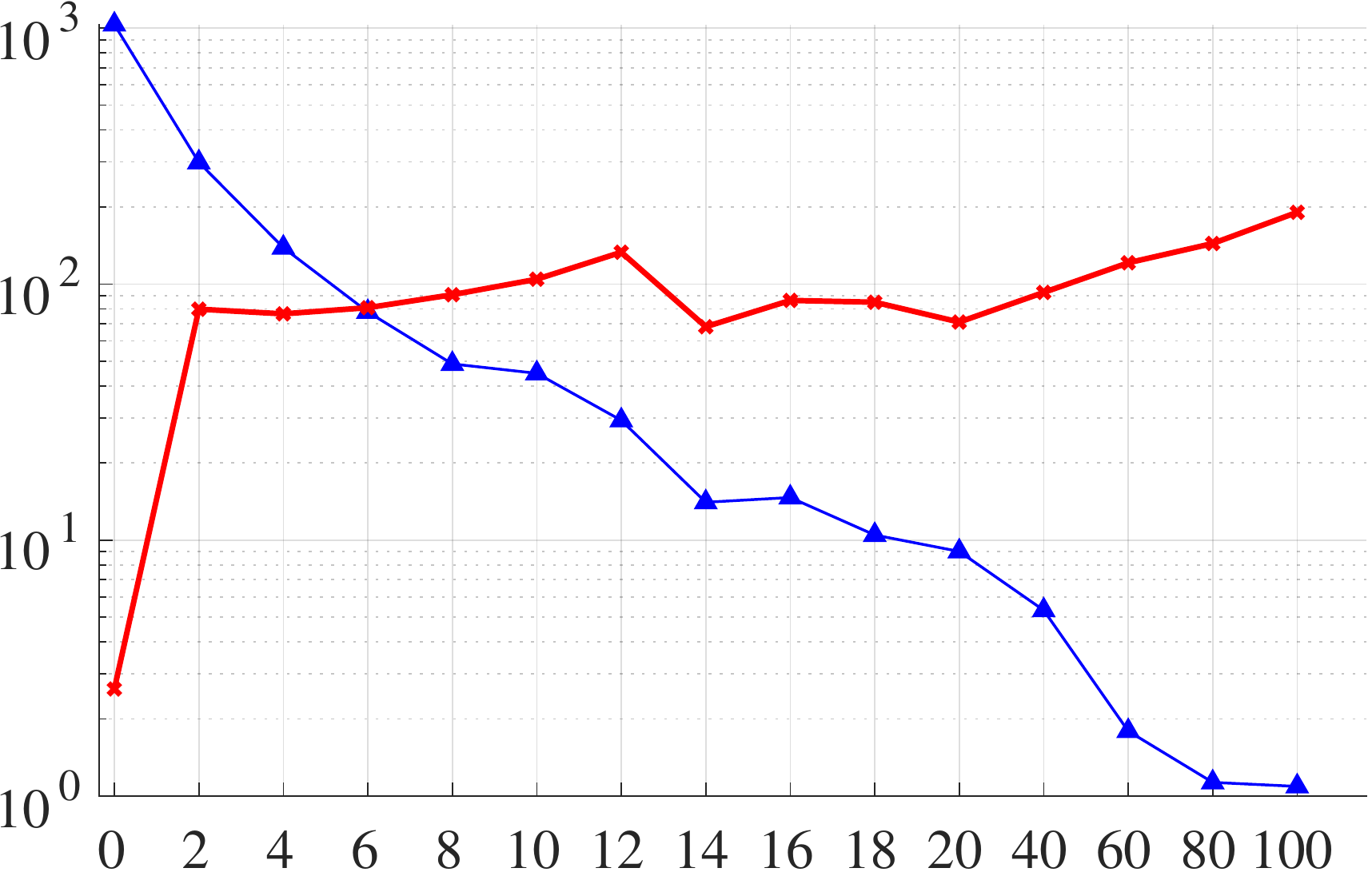} \\[.2em]
		& & \includegraphics[width=0.16\linewidth]{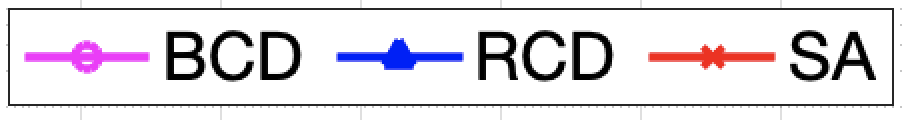} &\\[-.2em]
	\end{tabularx}	
	\caption{Runtime [s] ($y$-axis in log-scale) at varying graph densities $d_{G}$ ($x$-axis in $\times10^{-2}$) for SfM and SLAM graphs with $n=100, 200, 1000$ cameras. We denser sampled the interval $[0, 0.2]$.} 
% 	\caption{Runtimes (seconds in log-scale) at varying graph densities $d_{G}$ for SfM and SLAM aimed camera graphs with $n=100, 200, 1000$ cameras. We recorded runtimes for RCD, BCD, and SA. We denser sampled the interval $[0, 0.1]$ to 0.02; \textit{x-axis} is the graph densities $d_{\mathcal{G}}$; \textit{y-axis} is the time (seconds) in log-scale. For visibility, we only show subset of the intervals.} 
    \label{fig:synth}
\end{figure*}

\vspace{-3em}
\paragraph{Graph density}
Consider a connected graph $\cG = (\cV, \cE)$ with $n=|\cV|$ vertices and $m=|\cE|$ edges. Define
\begin{align}\label{eq:dg}
    %{d}_{\cG} := \frac{|\cE|-|\cE_{\min}|}{|\mathcal{E}_{\max}|-|\cE_{\min}|}\,,
    {d}_{\cG} := ({|\cE|-|\cE_{\min}|})/({|\mathcal{E}_{\max}|-|\cE_{\min}|})\,,
\end{align}
as the density of graph $\cG$, where $\cE_{\max}$ and $\cE_{\min}$ denote the set of edges of the complete $(\cV, \cE_{\max})$ and the cycle graph $(\cV, \cE_{\min})$. Excluding graphs with $n-1$ edges\footnote{Rotation averaging instances are typically overdetermined, i.e., problems with $|\cE| > n-1$ edges.}, $d_{G}$ takes values in $[0,1]$. Thus, by definition~\eqref{eq:dg},  $d_{G}=0$ for a cycle graph and $d_{G} =1$ for a complete graph. 

\subsection{Synthetic Data}\label{sec:synth}
% \subsubsection{Simulated SfM}
To test RCD over a variety of graph configurations, we synthesised graphs with varying densities to simulate SfM and SLAM problems. As SfM often solves reconstruction from views with large baselines, we generated random camera positions and random connections in the SfM setting. In contrast, for the SLAM setting, we simulated views with a smooth trajectory and connect only nearby views. We created measurements of relative rotations~\eqref{eq:relrot} by multiplying the ground truth relative rotations with rotations with random axes and angles normally distributed with $\sigma= 0.1$ rad. For a fair comparison, we initialised all methods with the same initial random absolute rotations.

\vspace{-1em}
\paragraph{Varying graph densities}

Fig.~\ref{fig:synth} shows the runtimes averaged over 10 runs for all methods. RCD significantly outperformed SA for $d_{\cG} > 0.1$. In general, camera graphs from real-world SfM datasets are often dense. See for example $d_{\cG}$ values for the real-world instances in Table~\ref{table:realworld} with average $d_\mathcal{G}\approx{0.53}$. For larger problems, BCD was not able to terminate within reasonable time ($\le 1000$s); we did not report results for BCD for $n>100$. Although RCD was not considerably faster than SA when $d_{\mathcal{G}} < 0.04$, the convergence rate can be accelerated by using a local optimisation routine as we show in Sec.~\ref{subsubsec:result_slam}.

 %($\approx{0.54}$)

% Fig.~\ref{fig:synth} shows the runtimes averaged over 50 runs for all methods. RCD significantly outperforms SA for $d_{\cG} > 0.1$. In general, we often deal with dense camera graphs in real world SLAM applications ($\approx{0.54}$) ; see column 4 in Table~\ref{table:realworld}. For larger problems where $n>100$, BCD is not able to terminate within reasonable time, so we did not report BCD for $n>100$.
% Although Algorithm \ref{alg:rcd} is not considerably faster than SA when $d_{\mathcal{G}} < 0.04$, the convergence rate can be accelerated by using a local optimisation routine; see Sec.~\ref{subsubsec:result_slam}.

\vspace{-1em}
\paragraph{Varying noise levels and number of cameras}

\begin{figure}
    \centering
    \hspace{-2mm}
    \subfigure[]{\includegraphics[width=0.46\linewidth]{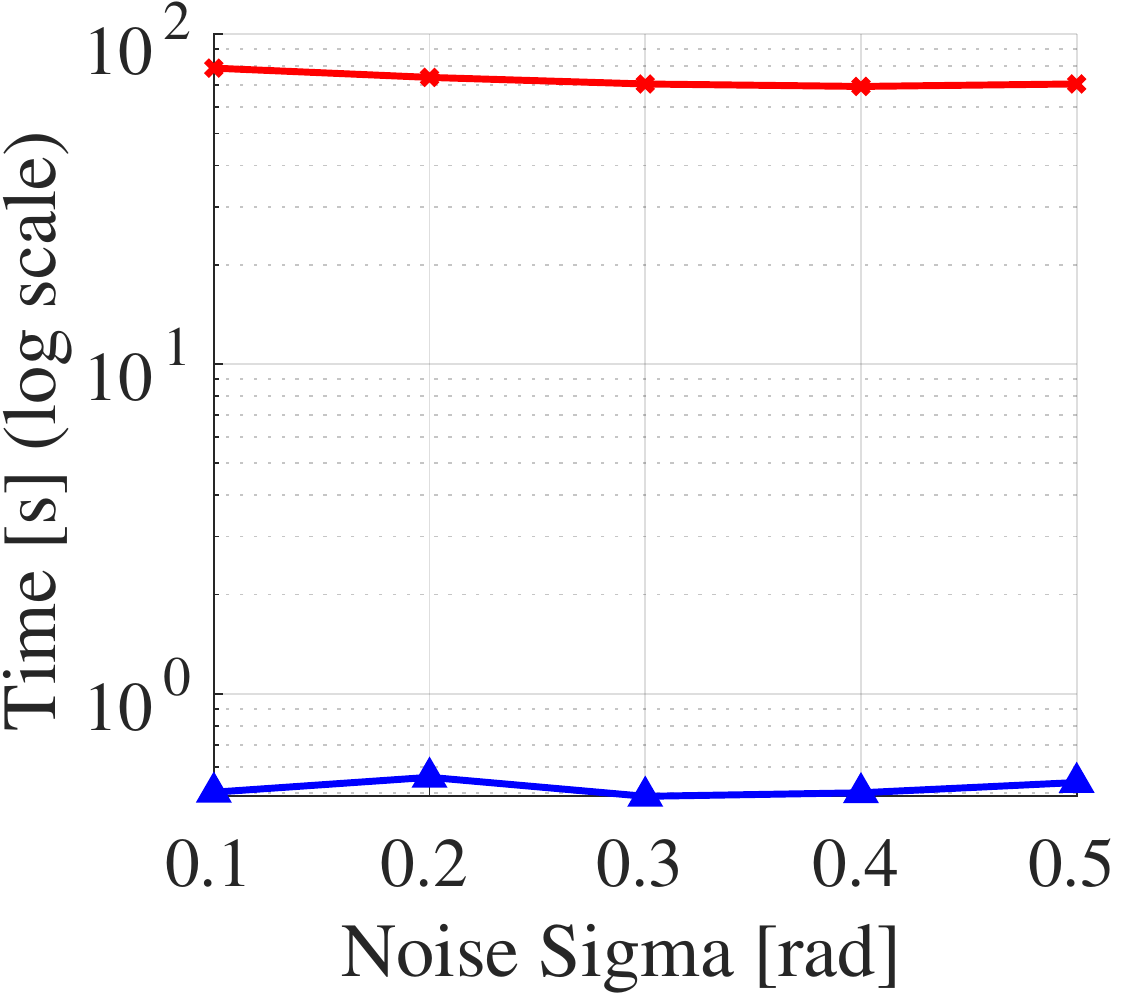} \label{fig:noise}}\;\;
    \subfigure[]{\includegraphics[width=0.45\linewidth, trim=0 0 0.5cm 0, clip]{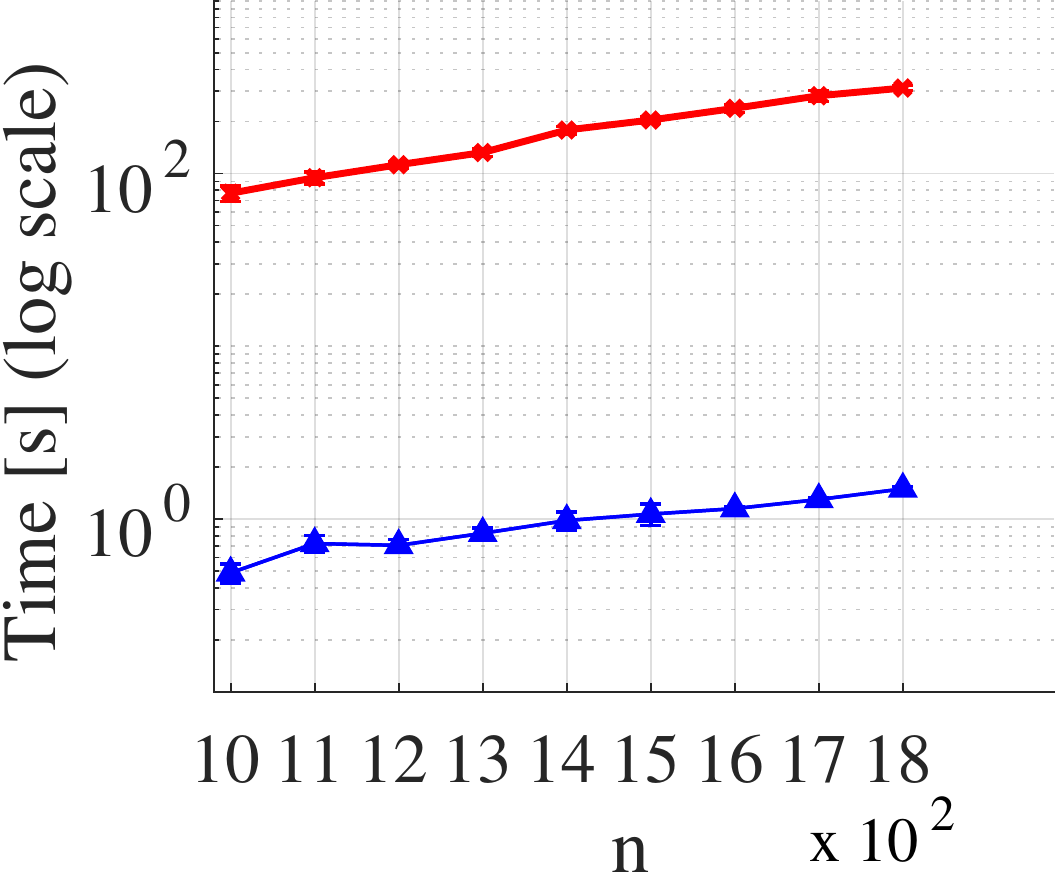}\label{fig:scala}}
    \caption{Runtime [s] (in log-scale) for SfM camera graphs with $d_\mathcal{G}=0.4$. (a) Varying $\sigma$ in $[0.1, 0.5]$ rad. and $n=1000$.  (b) Varying $n$ in $[1000, 1800]$ with $\sigma = 0.1$ rad. See Fig.~\ref{fig:synth} for the description of the legend.}
    % \caption{Runtimes (seconds in log-scale) at varying noise $\sigma$ and $n$ with $d_\mathcal{G}=0.4$ for SfM aimed camera graphs; (a) $\sigma = [0.1, 0.5]$rad with $n=1000$; (b) $n = [1000, 1800]$ with $\sigma = 0.1$rad.}
    \vspace{-1em}
    \label{fig:noise_and_scalability}
\end{figure}

In Fig.~\ref{fig:noise_and_scalability}, we plotted the runtimes of RCD and SA on SfM camera graphs with varying noise levels $\sigma$, number of cameras $n$, but with fixed $d_\mathcal{G}=0.4$. We omitted the comparison against BCD as it did not converge within a sensible time for large problems, as demonstrated in~Fig.~\ref{fig:synth}.  Fig.~\ref{fig:noise} shows that runtimes for RCD and SA were marginally affected by noise. Fig.~\ref{fig:scala} shows that RCD outperformed SA by two orders of magnitude ($1.5s$ vs $312.9s$ at $n=1,800$) --- this further demonstrates the superior scalability of RCD.

We repeated the above experiment with $d_\mathcal{G}=0.2$ which simulates SLAM graphs; see supp. material for the results.

% Slam benchmark result
\renewcommand{\tabcolsep}{2pt}
\begin{table*}[t]
    \centering
    \begin{tabular}{lrrr|cccccc|rrrrrrr}
      \toprule
        \multicolumn{4}{c|}{\textbf{Dataset characteristics}} &
        \multicolumn{1}{c}{} &
        \multicolumn{4}{c}{ \textbf{Error} [\%]}     &
        \multicolumn{1}{c|}{} &         
        \multicolumn{7}{c}{ \textbf{Efficiency}} \\
%          \noalign{\smallskip}    
        \multicolumn{1}{c}{} &
        \multicolumn{1}{c}{$|\mathcal{V}|$} &        
        \multicolumn{1}{c}{$|\mathcal{E}|$} &
        \multicolumn{1}{c|}{} &        
        \multicolumn{1}{c}{} &
        \multicolumn{1}{c}{} &
        \multicolumn{3}{c}{} & 
        \multicolumn{1}{c|}{} &         
        \multicolumn{2}{c}{\# Epoch} &
        \multicolumn{1}{c}{} &
        \multicolumn{3}{c}{Time [s]} &
        \multicolumn{1}{c}{Speedup}\\
        %\cmidrule{11-12} \cmidrule{14-16}     
        \multicolumn{1}{c}{Name}  & \multicolumn{1}{c}{$n$}  & \multicolumn{1}{c}{$m$}  &  \multicolumn{1}{c|}{$d_\mathcal{G}$}   &  &
        Init. & RCD & RCDL & SA &  &  
        RCD  & RCDL  & & 
        RCD  & RCDL    & SA & \\ 
        \hline
        %\noalign{\smallskip}     
        \emph{smallgrid}
        & 125   & 297 & 0.02200 & 
        & -16.13  & 0 & -4.77E-09  & -8.38E-05 &
        & 46 & 10 &
        & 0.07 & \textbf{0.02} & 0.06 & 2.7 \\        
        %\noalign{\smallskip}  
        
        \emph{garage}
        & 1661   & 6275 & 0.00340 &  
        &  -7.29E-05 & -3.63E-06 & 0  & -1.40E-07 & 
        & 29 & 2 &
        & 3.74 & \textbf{0.28} & 4.76 & 17.1\\
        %\noalign{\smallskip} 
        
        \emph{sphere}
        & 2500   &   4949   &   0.00078 & 
        & -1.70 & -6.24E-06    &   0   &  -7.84E-07 & 
        &  352  & 2 & 
        & 105.70  &    \textbf{0.66} &  17.07 & 25.7\\ 
        %\noalign{\smallskip} 
        %\hl{cubicle?}         
        %& 5750  & 16869    &0.00067  
        %& -9.91E-06  & 0   &  -3.70E-04
        %& 7 & 2 
        %& \textbf{0.11} & 0.60 & 10.64\\       
       
        \emph{torus}
        & 5000  & 9898 &   0.00039 & 
        & -20.95  & -2.55E-05 & 0    &   -1.73E-06 &      
        & 1620  & 4  &
        & 1808.86 & \textbf{4.86} & 15.76 & 3.2\\
        %\noalign{\smallskip}   
        
        \emph{grid3D}
        & 8000  & 22819  & 0.00046 &  
        & -15.41 & -4.54E-06  & 0 &  -2.14E-06 & 
        & 409 & 4  &
        & 1199.30 & \textbf{14.78} & 23.93 & 1.6\\
        %\noalign{\smallskip}  
        
    \bottomrule    
    \end{tabular}
    \caption{Quantitative results for the SLAM Benchmark dataset~\cite{carlone15}. Error of the initial solution (Init.) and each method is the $\%$ of its objective  w.r.t. the lowest obtained objective among all methods. One epoch is equivalent to $n$ iterations as described in Sec.~\ref{sec:speedup}. Speedup is presented for the best result of RCD and RCDL against SA.}
    %\caption{Comparison of computational efficiency and resulting errors on SLAM Benchmark dataset~\cite{carlone15}. 'Error (in $\%$)' is the error relative to the lowest objective value; 'One epoch' is $n$ iterations as described in Sec.~\ref{sec:speedup}; 'Speedup' is the speedup order of fastest proposed variant (RCD and RCDL), compared to SA.}
    \label{table:Slam_benchmark_result}
\end{table*}

\renewcommand{\tabcolsep}{2.5pt}
\begin{table*}
    \centering
    \begin{tabular}{lrrr|lccccc|rrrrrrrr}
      \toprule
        \multicolumn{4}{c|}{\textbf{Dataset characteristics}} &
        \multicolumn{1}{c}{} &
        \multicolumn{4}{c}{ \textbf{Error} [\%]}     &
        \multicolumn{1}{c|}{} &
        \multicolumn{8}{c}{ \textbf{Efficiency}} \\
        \multicolumn{1}{c}{} &
        \multicolumn{1}{c}{$|\mathcal{V}|$} &        
        \multicolumn{1}{c}{$|\mathcal{E}|$} &
        \multicolumn{1}{c|}{} &
        \multicolumn{1}{c}{} &        
        \multicolumn{4}{c}{} & 
        \multicolumn{1}{c|}{} &        
        \multicolumn{2}{c}{\# Epoch} & 
        \multicolumn{1}{c}{} &
        \multicolumn{3}{c}{Time [s]} &
        \multicolumn{1}{c}{} &
        \multicolumn{1}{c}{Speedup}\\
        %\cmidrule{11-12} \cmidrule{14-16} 
        
        \multicolumn{1}{c}{Name}   & \multicolumn{1}{c}{$n$}   & \multicolumn{1}{c}{$m$}  & \multicolumn{1}{c|}{$d_\mathcal{G}$} &  & Init. & RCD & RCDL & SA & 
        & RCD  & RCDL  &  & RCD  & RCDL  & SA & &  \\ 
        \hline
        %\noalign{\smallskip}
        \emph{Alcatraz Tower}
        & 172   & 14706 & 1.00  &
        & -0.66  & -8.66E-10 & 0  & -4.64E-08 & 
        & 4 & 2 &
        & \textbf{0.03} & 0.12 & 0.63 & & 25.3\\
         %\noalign{\smallskip}
        
        \emph{Doge Palace}             
        & 241   &   19753   &   0.68    & 
        & -0.89 &-1.16E-08    &   0   &  -1.04E-07 &     
        &  8  & 2 & 
        &\textbf{0.09}   &   0.21 &  1.00 & & 11.1\\ 
        %\noalign{\smallskip}
        
        \emph{King's College}          
        & 328  & 41995    & 0.78    &  
        & -1.57  & -5.01E-10   & 0   &  -3.81E-08 & 
        & 7 & 2 &
        & \textbf{0.11} & 0.60 & 2.37 & & 21.5\\  
        %\noalign{\smallskip}    
       
        \emph{Alcatraz Garden}    
        & 419  & 51635 &   0.59 &
        & -1.29  &   -7.70E-09 & 0    &   -5.57E-08 &        
        & 11  & 2   &  
        & \textbf{0.24} & 0.89 & 3.24 & & 13.5\\
        %\noalign{\smallskip}
        
        \emph{Linkoping}   
        & 538  & 34462  & 0.24  &  
        & -1.22  & -3.62E-07   & 0 &  -4.03E-06 &
        & 37 & 2    &   
        & 0.90 & \textbf{0.43} & 6.44 & & 15.0\\  
        %\noalign{\smallskip}    
        
        \emph{UWO}   
        & 692  & 80301    &  0.33   & 
        & -1.26  & -1.38E-07   & 0   &  -7.88E-07 &
        &   20  & 2 &
        &   \textbf{0.85} & 1.65 & 11.12 & & 13.1\\
        %\noalign{\smallskip}    
        
        \emph{Orebro Castle}   
        & 761  &  116589   & 0.40   &
        & -1.19  & -9.40E-08  &  0  & -1.04E-06 &
        &  20 &  2  &
        & \textbf{1.10}  & 4.01  & 26.17 & & 23.8\\ 
        %\noalign{\smallskip}    
        
        \emph{Spilled Blood}   
        & 781  &  117814  & 0.39    &   
        & -2.81  & -3.20E-08 & 0 &  -6.61E-07 &
        &  14   & 2 &
        & \textbf{0.79}  & 4.32  & 37.64 & & 47.6\\
        %\noalign{\smallskip}        
        
        \emph{Lund Cathedral}   
        & 1207  & 177289  & 0.24    &
        & -1.16 & -9.62E-07  & 0  & -1.91E-06 &   
        & 78  & 2   & 
        & 8.45 & \textbf{7.03} & 41.08 & & 5.8\\
        %\noalign{\smallskip}    
        
        \emph{San Marco}   
        & 1498  & 757037    & 0.67  &  
        & -0.74 & -6.60E-09  &  0  &  -8.97E-09 &
        &  6 &  2   &
        & \textbf{1.61} & 145.46 & 110.07 & & 68.4\\
        %\noalign{\smallskip}
        
        %Bascilica   & 1805  &     &   &     &    &  &            &  \\        
    \bottomrule    
    \end{tabular}
    \caption{Quantitative results for the SfM large scale real-world dataset~\cite{olsson11}. See Table~\ref{table:Slam_benchmark_result} for the description of each column.}
    \label{table:realworld}
\end{table*}

\subsection{SLAM benchmark dataset}\label{subsubsec:result_slam}

We compared runtimes on large-scale problems from the SLAM dataset in~\cite{carlone15}. Table~\ref{table:Slam_benchmark_result} reports the input characteristics of each benchmarking instance and the results for all methods. Here, we initialised all algorithms with the same initial rotations from a random spanning tree. Note that initialisation does not affect the global optimality of tested algorithms. The spanning tree initialisation is fast and practical. We remark that in real-world applications it is unnecessary to solve camera orientations from random rotations. %%Again, we initialised all algorithms with the same initial estimates for fair comparison. 
We provided the errors (in $\%$) of resulting objective (including the initialisation) relative to the lowest objective value reported among all methods.

Camera graphs are very sparse for the SLAM Benchmark in Table~\ref{table:Slam_benchmark_result} ($d_\mathcal{G} \le 0.022$). Although RCD was not as fast as SA on \emph{sphere}, \emph{torus} and \emph{grid3D} (note that $d_\mathcal{G} \le 0.0007$ in those instances), the use of a local optimisation in RCDL permitted to outperform SA; see also Fig.~\ref{fig:slam_RCD_vs_RCDL}.

% To demonstrate \hl{the effect of local optimisation on (check..)} the convergence of RCDL, Fig.~\ref{fig:slam_RCD_vs_RCDL} plots the objective value for RCD and RCDL at increasing epochs. During the 1st epoch, the local algorithm drastically reduced the objective (from point in \textit{green} to point in \textit{magenta}). This reveals the collaborative strength of global and local solvers, which enables RCDL to transverse through the $SO(3)$ more efficiently and thus terminate at the much lesser epochs (\textit{red dot}) compared to RCD (\textit{blue dot}). \hl{say something about we could use better local mehtods, lesser number of iterations, etc}
%\hlgreen{due to the sparsity of the graph integrating a local optimisation scheme (RCDL) has accelerated the convergence.} 

%It is evident that the integration of the local optimisation scheme has assisted Algorithm~\ref{alg:rcd} to transverse through $SO(3)$ much more efficiently and converge to the globally optimal solutions faster than RCD and SA. 

%\paragraph{Accelerating convergence with local optimisation}
%\begin{itemize}
%    \item Why RCDL is slower than the RCD in Table 2.
%    \item Although the time taken is longer, the number of epoch has been significantly reduced.         90\% of the computational time is spent on solving the local algorithm. 
%    \item Perhaps? Do another plot (like torus) for San Marco dataset?
%\end{itemize}

\subsection{Real-world SfM dataset}
\begin{figure}
\vspace{-1mm}
    \centering
    \includegraphics[width=.8\linewidth]{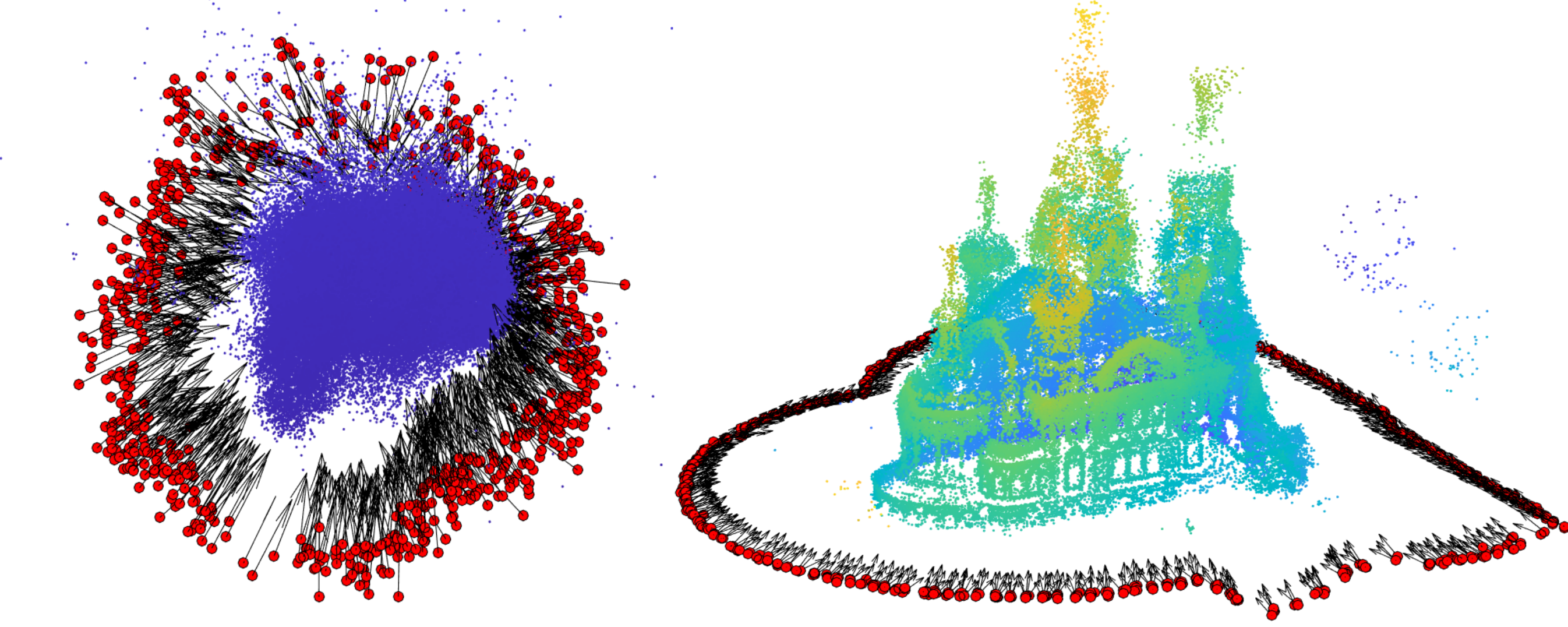}
    \caption{Reconstruction of the \textit{Spilled Blood Cathedral} by solving the \emph{known-rotation-problem}~\cite{zhang2018fast}. \emph{Left}: Initial camera orientations. \emph{Right}: Result from RCDL after $1$ epoch.}
    \vspace{-1em}
    \label{fig:sfm_RCD_vs_RCDL}
\end{figure}

Table~\ref{table:realworld} presents runtimes over real-world SfM datasets~\cite{olsson11}
%\footnote{http://www.maths.lth.se/matematiklth/personal/calle/dataset/dataset.html} 
where RCD outperformed SA; see the supp. material for errors in $\degree$. We remark that RCDL took substantially fewer epochs compared to RCD to converge. However, RCDL did not achieve a better runtime as local optimisation consumed on average $\approx 90\%$ of the total runtime, especially for large graph densities. Fig.~\ref{fig:sfm_RCD_vs_RCDL} shows the reconstructed~\emph{Spilled Blood} using the estimated camera orientations of RCDL after running for \textit{one} epoch in $4.2$s. 

%Although the computational time of RCDL on some of the sequences are higher than RCD, the local optimisation scheme substantially reduces the number of epochs. Note that most of the total runtime in RCDL is on solving local optimisation. For e.g., $97\%$ of the computational time ($\approx{4.2\text(s)}$) is spent by local solver. We compare the angular error of the estimated rotations of both RCD and RCDL after $1$ epoch. The maximum angular error of RCDL estimation is $0.92\degree$, compared to the RCD estimation which has $12.3\degree$. 
%\hl{say something we could chosen a faster method..}
%Fig.~\ref{fig:sfm_RCD_vs_RCDL} depicts the reconstructed \textit{Spilled Blood Cathedral} after solving a known rotation problem~\cite{zhang2018fast} using the estimated camera orientations of RCDL after $1$ epoch. $97\%$ ($4.2\text{s})$ of the computational time is contributed by the local solver -- the maximum reprojection error of $17.4$ px., compared to the one using the estimated camera orientations of RCD after $1$ epoch with 34.96 px. \hl{improve this paragraph, make clear its goal...}

\section*{Acknowledgements}
This research was supported by the Australian Research Council through the grants ARC DP200101675, ARC FT170100072, and the ARC Centre of Excellence for Robotic Vision CE140100016.

%\clearpage
\section{Conclusions}
We present RCD, a fast rotation averaging algorithm that finds the globally optimal rotations under mild conditions on the noise level of the measurements.  
Our insights on gauge freedom has circumvented the quadratic computational burden of BCD, which is an established method for global rotation averaging. Also, since RCD maintains valid rotations instead of a dense PSD matrix, local optimisation routines can be beneficially integrated. Experimental results demonstrated the superior efficiency of RCD, which significantly outperformed state-of-the-art algorithms on a variety of problem configurations.

% v2
%We present RCD, a fast rotation averaging algorithm that finds the globally optimal rotations under mild conditions on the noise level of the measurements. Despite RCD is inspired by BCD, RCD is intrinsically superior. 
%Our insights on a special case of gauge freedom has circumvented the quadratic computational burden of BCD which is an impediment for practical applications.
%Also, maintaining valid rotations instead of a dense PSD matrix facilitates the integration of local optimisation routines.
%Experimental results demonstrate the superior efficiency of RCD, significantly outperforming Shonan, on a variety of problem configurations.

% v1
%We presented RCD, a fast rotation averaging algorithm that finds the globally optimal rotations under mild conditions on the noise level of the measurements. Despite RCD was inspired in BCD, RCD is intrinsically superior as it avoids the quadratic burden of BCD on the size of the solution. Also, by updating rotations instead of a large PSD matrix, RCD facilitates the incorporation of local optimisation routines. 
%Experimental results demonstrated the superiority of RCD over Shonan, the state-of-the-art global method, on a variety of problem configurations.

%needing to estimate much more than $100$ orientations

\clearpage
{\small
\bibliographystyle{ieee_fullname}
\bibliography{rotavg}
}

\end{document}

%%%%%%%%% TITLE
\title{Supplementary Material for: \\Rotation Coordinate Descent for Fast Globally Optimal Rotation Averaging}

\title{Rotation Coordinate Descent for Fast Globally Optimal Rotation Averaging}
\author{Álvaro Parra$^{1\thanks{equal contribution}}$ \hspace{1em} Shin-Fang Chng$^{1 \footnotemark[1]}$ \hspace{1em} Tat-Jun Chin$^{1}$ \hspace{1em} Anders Eriksson$^{2}$ \hspace{1em}  Ian Reid$^{1}$  \\
$^{1}$ School of Computer Science, The University of Adelaide \\
$^{2}$ School of Information Technology and Electrical Engineering, University of Queensland}

\maketitle

%%%%%%%%% BODY TEXT
%-------------------------------------------------------------------------

\section{Additional Results}

\subsection{Varying noise levels and number of cameras in SLAM graphs}
\begin{figure}[h]
    \centering
    \hspace{-2mm}
    \subfigure[]{\includegraphics[width=0.49\linewidth]{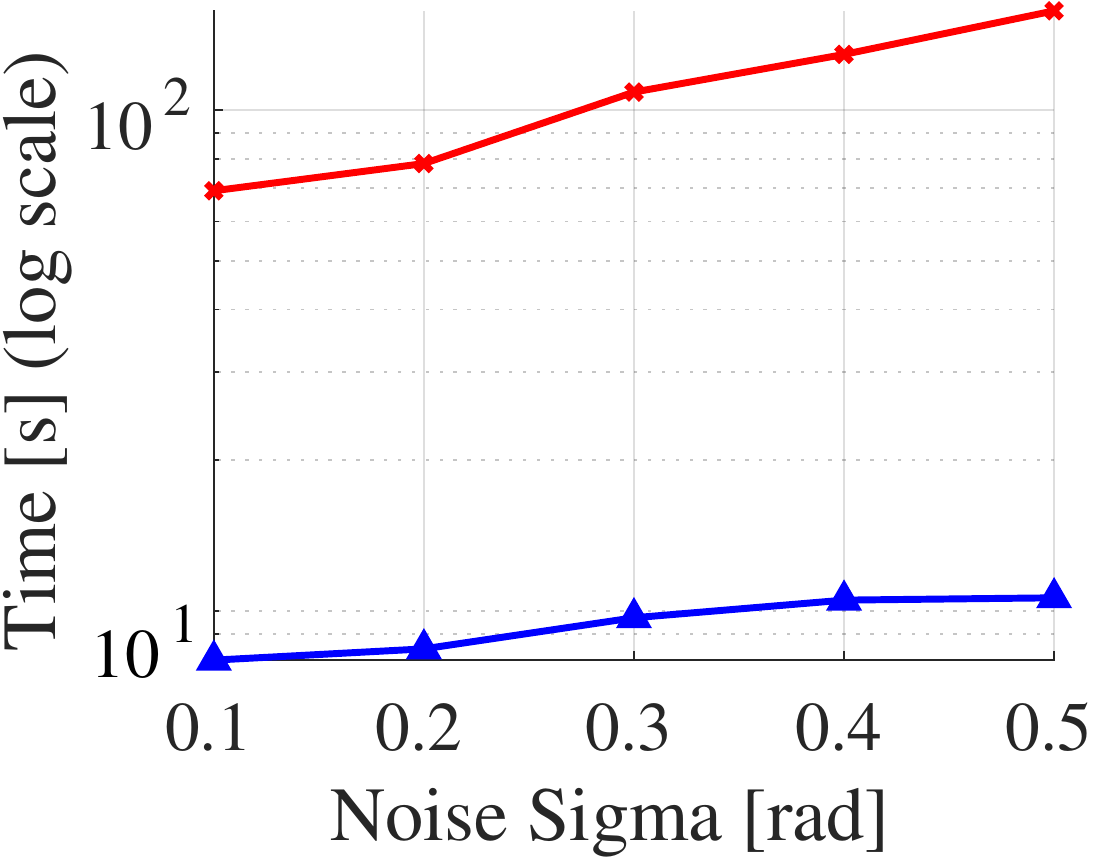} \label{fig:noise}}
    \subfigure[]{\includegraphics[width=0.49\linewidth, trim=0 0 0.5cm 0, clip]{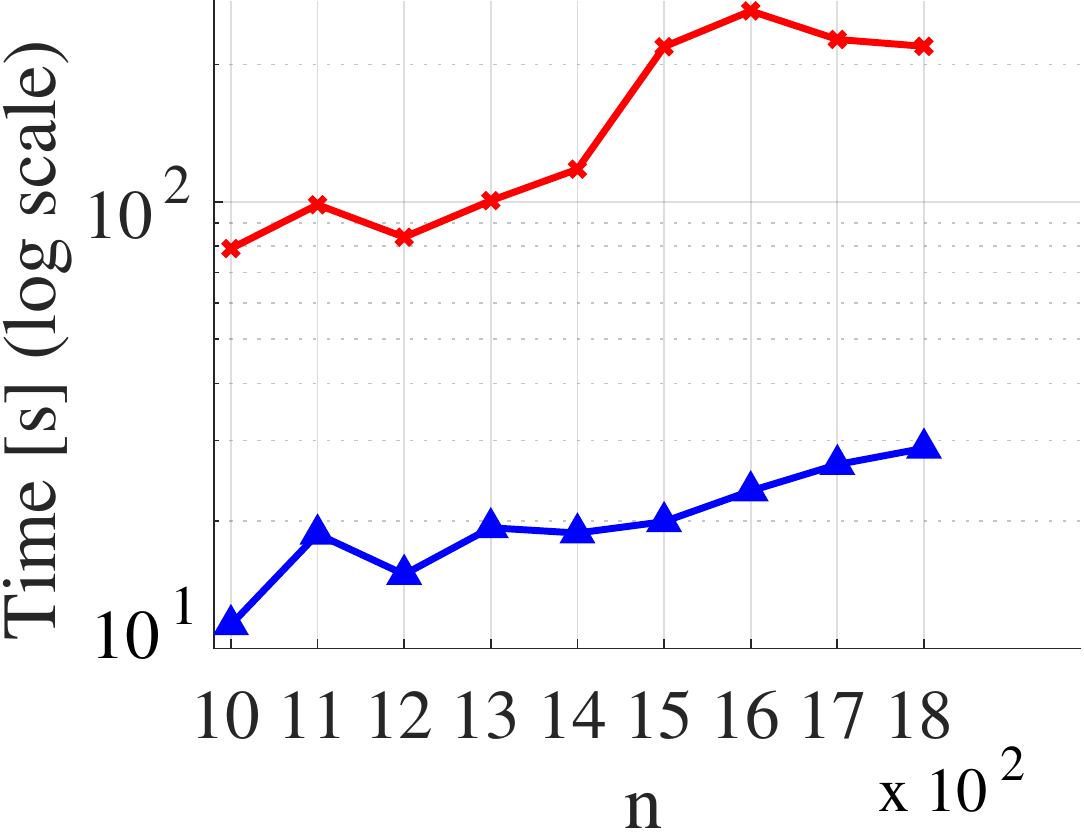}\label{fig:scala}}
    \caption{Runtime [s] (in log-scale) for SLAM camera graphs with $d_\mathcal{G}=0.2$. (a) Varying $\sigma$ in $[0.1, 0.5]$ rad. and $n=1000$.  (b) Varying $n$ in $[1000, 1800]$ with $\sigma = 0.1$ rad. See Fig.~3 in the main text for the description of the legend.}
    % \caption{Runtimes (seconds in log-scale) at varying noise $\sigma$ and $n$ with $d_\mathcal{G}=0.4$ for SfM aimed camera graphs; (a) $\sigma = [0.1, 0.5]$rad with $n=1000$; (b) $n = [1000, 1800]$ with $\sigma = 0.1$rad.}
    \label{fig:noise_and_scalability}
\end{figure}

\subsection{Quantitative results for the SfM large scale real-world dataset (Error in $\degree$)}
See Table \ref{table:realworld}.
\begin{table*}[h]
    \centering
    \begin{tabular}{lrrr|c|c|c|c|c|c}
      \toprule
        \multicolumn{4}{c|}{\textbf{Dataset characteristics}} &
         \multicolumn{6}{c}{ Error [$^{\circ}$]}\\ 
        %\cmidrule{11-12} \cmidrule{14-16} 
        
        \multicolumn{1}{c}{Name}   & \multicolumn{1}{c}{$n$}   & \multicolumn{1}{c}{$m$}  & \multicolumn{1}{c|}{$d_\mathcal{G}$} & \multicolumn{2}{c|}{Mean} & \multicolumn{2}{c|}{Median} & \multicolumn{2}{c}{Max}\\
        & & & & Init. & RCD  & Init. & RCD  & Init. & RCD \\ 
        \hline
        %\noalign{\smallskip}
        \emph{Alcatraz Tower}          
        & 172   & 14706 & 1.00  
        & 4.65 & 0.63  & 4.16 & 0.63  & 20.57 & 0.63 \\
        
        \emph{Doge Palace}             
        & 241   &   19753   &   0.68    
        & 5.95 & 0.56 & 5.18 &  0.52 & 23.07 & 1.82 \\ 
        %\noalign{\smallskip}
        
        \emph{King's College}          
        & 328  & 41995    & 0.78   
        &  6.98 & 0.39  & 4.89   & 0.36   &  32.99 & 0.83 \\  
        %\noalign{\smallskip}    
       
        \emph{Alcatraz Garden}    
        & 419  & 51635 &   0.59 
        & 7.79  &  0.92 & 6.62 & 0.92 & 25.65 & 1.79 \\
        %\noalign{\smallskip}
        
        \emph{Linkoping}   
        & 538  & 34462  & 0.24  
        &  7.42 & 0.63  & 6.90   & 0.58 & 25.15 & 2.63 \\  
        %\noalign{\smallskip}    
        
        \emph{UWO}   
        & 692  & 80301    &  0.33   
        & 7.20 & 0.55  & 6.42   & 0.52 & 24.43  &  2.19  \\
        %\noalign{\smallskip}    
        
        \emph{Orebro Castle}   
        & 761  &  116589   & 0.40   
        & 6.66 & 0.52  & 5.55  &  0.50  & 27.30 & 1.33 \\ 
        %\noalign{\smallskip}    
        
        \emph{Spilled Blood}   
        & 781  &  117814  & 0.39    
        &  11.41 & 6.81  & 7.61 & 6.46 & 42.82 & 12.35 \\
        %\noalign{\smallskip}        
        
        \emph{Lund Cathedral}   
        & 1207  & 177289  & 0.24    
        & 8.05 & 0.40  & 7.52  & 0.34  & 27.12 & 3.52 \\
        %\noalign{\smallskip}    
        
        \emph{San Marco}   
        & 1498  & 757037    & 0.67  
        &  5.03 & 0.22 & 4.45  &  0.21  & 26.59 & 0.94 \\
        %\noalign{\smallskip}
        
        %Bascilica   & 1805  &     &   &     &    &  &            &  \\        
    \bottomrule    
    \end{tabular}
    \caption{Quantitative results for the SfM large scale real-world dataset~[25]. See Table~2 in the main text for the description of each column.}
    \label{table:realworld}
\end{table*}

%\section{Demonstration program}
%To run the demonstration program, please follow the instructions in the \verb|README.md| file in folder \verb|demo_RCD|.

\section{Further details}
\subsection{Conditions on the noise level for the strong duality of Eq. (5)} \label{sec:sd} %(L132)

For the following rotation averaging problem (Eq. (5) in the main text)
\begin{align}\label{eq:special}
\min_{R_1, \ldots, R_n \in SO(3)} \sum_{(i,j) \in \cE } d_\text{chordal}(R_j R_i^T, \tilde{R}_{ij})^2 ,
\end{align}
we present a bound on the \emph{angular} residual errors 
\begin{align}
\alpha_{ij} = d_{\angle} (R_j^* {R_i^*}^T, \tilde{R}_{ij})
\end{align}
such that its \emph{strong duality} holds.

% Here we present a bound on the \emph{angular} residuals errors 
% \begin{align}
% \alpha_{ij} = d_{\angle} (R_j^* {R_i^*}^T, \tilde{R}_{ij})
% \end{align}
% for the \emph{strong duality} of the following rotation averaging problem (Eq. (5) in the main text)
% \begin{align}\label{eq:special}
% \min_{R_1, \ldots, R_n \in SO(3)} \sum_{(i,j) \in \cE } d_\text{chordal}(R_j R_i^T, \tilde{R}_{ij})^2 .
% \end{align}

The main result of~[Theorem 4.1, 11] is the proof of the strong duality of Problem~\eqref{eq:special} if 
\begin{align}
| \alpha_{ij} | \leq \alpha_{\max} \quad \forall (i,j) \in \cE, 
\end{align}
where
\begin{align} \label{eq:alpha_max}
\alpha_{\max} = 2 \arcsin{\left(\sqrt{\dfrac{1}{4} + \dfrac{\lambda_2 (L_G) }{2 d_{\max}}}-\dfrac{1}{2}\right)} . 
\end{align}
$\lambda_2(L_G)$ and $d_{\max}$ in~\eqref{eq:alpha_max} are related to the structure of the camera graph. More precisely, $\alpha_{\max}$ depends on the connectivity of the camera graph represented by its Fiedler value $\lambda_2(L_G)$ (the second smallest eigenvalue of its Laplacian $L_G$), and its maximal vertex degree $d_{\max}$ (c.f. to~[11] and~[12] for more details).

From the dependency of $\alpha_{\max}$ on the structure of the camera graph, it can be established that the most favourable case (admitting the largest residuals) is the complete graph for which $\alpha_{\max} \approx 42.9\degree$. The other extreme case is a cycle with $\alpha_{\max} = \pi/n$,  which induces a low angular bound for a large number of cameras although [11] suggested that this bound was ``quite conservative''. 

Although conditions were presented in terms of the angular distance, we remark that a chordal bound can also be established for the chordal residuals $\{d_\text{chordal}(R_j^* {R_i^*}^T, \tilde{R}_{ij})\}$ of Problem~\ref{eq:special} as both distances are related~[16]:
\begin{align}\label{eq:eq}
d_{\text{chordal}} (R, S) =  2 \sqrt{2}\sin\left(\dfrac{d_{\angle} (R,S)}{2}\right).
\end{align}

%$\lambda_2(L_G)$ and $d_{\max}$ in~\eqref{eq:alpha_max} are related to the structure of the camera graph. More precisely, $\alpha_{\max}$ depends on the connectivity of the camera graph represented by its Fiedler value $\lambda_2(L_G)$ (the second smallest eigenvalue of its Laplacian $L_G$), and its maximal vertex degree $d_{\max}$ (c.f. to~\cite{eriksson18} and~\cite{eriksson19} for more details). For example, in the case of a complete graph, the strong duality holds if the maximum residual is lesser than $\alpha_{\max} \approx 42.9\degree$; whereas for the extreme case of a cycle, $\alpha_{\max} = \pi/n$ it is not so favourable (for a very large $n$) although~\cite{eriksson18} claimed that this is a ``quite conservative'' for which ``it is possible to show a much stronger upper bound using a different analysis''. 

\subsection{Zero duality gap between (P) and (DD) } %(L291)

%\subsection{Proof of Eq.~\ref{eq:sd} in Section~\ref{subsec:duality} and Section~\ref{sec:sdp}}

Eriksson \etal~[11] have proven that under mild conditions on the noise level (see Sec.~\ref{sec:sd}), there is zero duality gap between their primal problem (P$_\text{orig}$) and their SDP relaxation (DD$_\text{orig}$). Since we defined our primal problem (P) and its SDP relaxation (DD) following a different convention for the relative rotation definition than~[11], here we show that our (P) and (DD) problems are equivalent to their counterparts in~[11]. Hence the zero duality gap extends to them. 

We defined our primal problem as follows. By rewriting the chordal distance using trace, \eqref{eq:special} becomes (Eq.~(8) in the main text)
\begin{align}\label{eq:rotavg_1}
 	 \min_{R_1, \ldots, R_n \in SO(3)}  -\sum_{(i,j)\in \cE}\tr(R_j^T \tilde{R}_{ij} R_i).
\end{align}
By the transpose invariance of the trace, \eqref{eq:rotavg_1} is equivalent to
\begin{align}\label{eq:rotavg_2}
 	 \min_{R_1, \ldots, R_n \in SO(3)}  -\sum_{(i,j)\in \cE}\tr(R_i^T \tilde{R}_{ij}^T R_j).
\end{align}
Our primal definition comes from rewriting~\eqref{eq:rotavg_1} more compactly as
\begin{align}\label{eq:p}
\tag{P}
\min_{R\in SO(3)^n} - \tr(R^T \tilde{R} R) 
\end{align}
using matrix notations, where
\begin{align}\label{eq:abs}
%R = [R_1\, R_2 \cdots R_n] \in SO(3)^n,
R = \left[R_1^T\, R_2^T \cdots R_n^T\right]^T \in SO(3)^n
\end{align}
contains the target variables, and $\tilde{R}$ encodes the transposes of the relative rotations. $\tilde{R}$ is then defined as
\begin{align}
    \tilde{R} = \begin{bmatrix}
    0_3 & a_{12}R_{12}^T & \cdots & a_{1n}R_{1n}^T\\
    a_{21}R_{21}^T & 0_3   & \cdots & a_{2n}R_{2n}^T\\
    \vdots & 0_3   & \ddots & \vdots\\
    a_{n1}R_{n2}^T &  a_{n2}R_{n2}^T & \cdots & 0_3\\
    \end{bmatrix},
\end{align}
where $a_{ij}$ are the elements of the adjacency matrix $A$ of $\cG$.

We now show that~\eqref{eq:p} is equivalent to the primal in~[11], which is defined as (Eq. (11) in~[11])
\begin{align}\label{eq:p_orig}
\tag{$\text{P}_\text{orig}$}
\min_{Q\in SO(3)^n} - \tr(Q \tilde{Q} Q^T) ,
\end{align}
where $Q$ is a ``row" vector containing rotation matrices
\begin{align}
Q = \left[Q_1, \ldots, Q_n \right],
\end{align}
and $\tilde{Q}$ encodes the relative measurements as 
\begin{align}
    \tilde{Q} = \begin{bmatrix}
    0_3 & a_{12}Q_{12} & \cdots & a_{1n}Q_{1n}\\
    a_{21}Q_{21} & 0_3   & \cdots & a_{2n}Q_{2n}\\
    \vdots & 0_3   & \ddots & \vdots\\
    a_{n1}Q_{n2} &  a_{n2}Q_{n2} & \cdots & 0_3\\
    \end{bmatrix} .
\end{align}
However, relative rotations ${Q}_{ij}$ in~[11] are defined such that (Eq. (4) in~[11])
\begin{align}
 {Q}_{ij} = Q_i^T Q_j. 
\end{align}
Contrast to our definition from Eq.~(1) in the main text where  we define relative rotations in the ideal case as
\begin{align}
    R_{ij} = R_j R_i^T. 
\end{align}
The following equivalences can then be established:
\begin{align}
    R_{i} = Q_i^T \text{ and } R_{ij} = Q_{ij}^T,
\end{align}
which implies that $Q = R^T$,  $\tilde{Q} = \tilde{R}$, and therefore \eqref{eq:p} is equivalent to \eqref{eq:p_orig} in the sense that their objective values are the same and their optimisers are related by a translation. 

Similarly, our SDP relaxation
\begin{subequations}
\begin{align}\label{eq:dd}
\tag{DD}
 & \min_{Y \in \mathbb{R}^{3n \times 3n}}&&  - \tr(\tilde{R}Y)\\
& \text{s.t.} &&  Y_{i,i} = I_3,\; i=1,\ldots,n. \label{eq:dd_diag}\\
&&& Y \succeq 0, \label{eq:dd:sdp}
\end{align}
\end{subequations}
is equivalent to its counterpart in~[11]. In effect, they are the same as matrices encoding rotations are the same for both problems ($\tilde{Q} = \tilde{R}$). 

%As we established that ours~\eqref{eq:p} and ~\eqref{eq:dd} problems are equivalent to the ones defined in~[10], we can extend the zero duality gap between (P) and (DD) in [10] to ours. We remark that the zero duality gap holds under mild conditions on the noise level (see~\ref{sec:sd}). 

\subsection{Validity of Algorithm~1 as equivalent to BCD in Eriksson et al.~[11] } %(L320)
% \subsection{Proof of Equivalence between Algorithm~\ref{alg:bcd} and original in Section~\ref{sec:gauge}}

Here we show that BCD as presented in Algorithm 1 in the main text is equivalent to the original BCD algorithm for rotation averaging proposed in~[11]. To facilitate presentation, we call BCD-Ours to Algorithm 1 in the main text and BCD-Orig to Algorithm~1 in~[11].

The improvement of BCD-Ours over BCD-Orig is that instead of creating a temporary large square matrix
\begin{align}
    B = \begin{bmatrix} 
Y_{(1:k-1);(1:k-1)}^{(t)} & Y_{(1:k-1);(k+1:n)}^{(t)} \\
Y_{(k+1:n);(1:k-1)}^{(t)} &  Y_{(k+1:n);(k+1:n)}^{(t)} \\
\end{bmatrix}
\end{align}
as in BCD-Orig, BCD-Ours creates a temporary vector which allows to operates directly on $Y^{(t)}$ as we will show next. 

Note that $B$ are the elements in $Y^{(t)}$ that are kept constant during the current iteration in BCD-Orig and BCD-Ours. On the other hand, the updated components for $Y^{(t)}$ in BCD-Orig are obtained from the optimiser $X^*$ of an SDP problem (Problem (26) in the main text) which has the following explicit solution:
\begin{align}\label{eq:closedform_orig}
    X^* = BC \left[ \left( C^T B C  \right)^\frac{1}{2} \right]^\dagger,
\end{align}
where $C \in \mathbb{R}^{3(n-1)\times 3}$ is the $k$-th column of $\tilde{R}$ without its $k$-th row, i.e.,
\begin{align}
    C = \begin{bmatrix}
    \tilde{R}^{(t)}_{(1:k-1);(k:k)}\\
    \tilde{R}^{(t)}_{(k+1:n);(k:k)}
    \end{bmatrix}.
\end{align}

Instead of computing the updates from~\eqref{eq:closedform_orig}, BCD-Ours solves
\begin{align}\label{eq:closedform}
    S = Z \left[ \left( W^T Z  \right)^\frac{1}{2} \right]^\dagger,
\end{align}
where $W \in \mathbb{R}^{3n \times 3}$ is the $k$-th column of $\tilde{R}$, i.e.,
\begin{align}
    W = \tilde{R}_{:,k} \,,
\end{align}
and 
\begin{align}
    Z = Y^{(t)} W
\end{align}
is a temporary vector.

We will show next that $X^*$ es equal to $S$ without its $k$-th element. Since BCD-Ours ignores the $k$-th element of $S$ during the update (Line 7 in BCD-Ours), BCD-Ours and BCD-Orig produce the same output. 

Note first that the pseudo-inverse parts  of~\eqref{eq:closedform_orig} and~\eqref{eq:closedform} are the same since
\begin{align}
     C^T B C = W^T Z  
\end{align}
as the $k$-th element in $W$ is zero ($W$ is the $k$-th column of $\tilde{R}$ which has diagonal elements equal to $0_3$). Similarly $BC$ is equal to $Z$ if removing the $k$-th element of $Z$. Hence~\eqref{eq:closedform} produces $X^*$ after removing the $k$-th element of $S$.

%To avoid creating auxiliary vectors in $\mathbb{R}^{3(n-1) \times 3}$ (e.g., $C$) to obtain $X^*$ directly following~\eqref{eq:closedform_orig}, Algorithm~\ref{alg:bcd} creates a vector $S \in \mathbb{R}^{3n \times 3}$ (Line~\ref{alg:bcd:s}) equal to $X^*$ if removing the $k$-th element in $S$. 

%Algorithm~\ref{alg:bcd} computes $S$ from the $\mathbb{R}^{3n \times 3}$ vectors $W$ and $Z = Y^{(t)} W$ (Lines~\ref{alg:bcd:w}~and~\ref{alg:bcd:z}) as:

% In Algorithm~\ref{alg:bcd} (Line~\ref{alg:bcd:z}), since~\eqref{eq:y0}
% \begin{align}
%     Z =  Y^{(0)} W = R^{(t)} ({R^{(t)}}^T W)
% \end{align}
% which is the value of $Z$ in Algorithm~\ref{alg:rcd} (Line~\ref{alg:rcd:z}).

%\subsection*{Improvements in Algorithm~\ref{alg:bcd} to BCD in~\cite{eriksson18}}
% From the definition of $R^{(1)}_{\text{BCD}}$ in~\eqref{eq:y0}, $R^{(1)}_{\text{BCD}}$ is the first column in $Y^{(1)}$~\eqref{eq:y1}, i.e., 
% \begin{align}
%  	R_{\text{BCD}}^{(1)} = \left[I_3\;\; {X^*}^T\right]^T,
% \end{align}

%-------------------------------------------------------------------------